\pdfoutput=1

\documentclass[11pt,usenames,dvipsnames]{article}

\usepackage{acl}
\usepackage{color}
\usepackage{colortbl}
\usepackage{times}
\usepackage{latexsym}
\usepackage{tabularx}
\usepackage{multirow}
\usepackage{booktabs} 
\usepackage{amsfonts}
\usepackage[T1]{fontenc}
\usepackage{adjustbox}
\usepackage{smiles}
\usepackage{bbm, dsfont}

\usepackage[utf8]{inputenc}
\usepackage[ruled,vlined]{algorithm2e}
\usepackage{microtype}

\newcommand{\ours}{\textsc{AcTune}}

\definecolor{light-gray}{gray}{0.9}

\title{{\ours}: Uncertainty-Based Active Self-Training for\\ Semi-Supervised Active Learning 
with Pretrained Language Models}

\author{Yue Yu$^1$, Lingkai Kong$^1$, Jieyu Zhang$^2$, Rongzhi Zhang$^1$, Chao Zhang$^1$ \\
        $^1$Georgia Institute of Technology, Atlanta, GA \ \ \ $^2$University of Washington, Seattle, WA  \\ \{yueyu, lkkong, rongzhi.zhang, chaozhang\}@gatech.edu, \ jieyuz2@cs.washington.edu}

\begin{document}
\maketitle
\begin{abstract}
  While pre-trained language model (PLM) fine-tuning has achieved strong
  performance in many NLP tasks, the fine-tuning stage can be still demanding
  in labeled data. Recent works have resorted to active fine-tuning to improve
  the label efficiency of PLM fine-tuning, but none of them investigate the
  potential of unlabeled data. We propose {\ours}, a new framework that
  leverages unlabeled data to improve the label efficiency of active
  PLM fine-tuning. \ours~ switches between data
  annotation and model self-training based on uncertainty: it selects
  high-uncertainty unlabeled samples for active annotation and low-uncertainty
  ones for model self-training. Under this framework, we design (1) a
  region-aware sampling strategy that reduces redundancy when actively
  querying for annotations and (2) a momentum-based memory bank that
  dynamically aggregates the model's pseudo labels to suppress label noise in
  self-training. Experiments on 6 text classification datasets show that
  {\ours} outperforms the strongest active learning and self-training
  baselines and improves the label efficiency of PLM fine-tuning by 56.2\% on
  average. {Our implementation will be available at
    \url{https://github.com/yueyu1030/actune}}.

\end{abstract}

\section{Introduction}
Fine-tuning pre-trained language models (PLMs) has achieved much success in
natural language processing
(NLP)~\citep{bert,liu2019roberta,brown2020language}. One benefit of PLM fine-tuning is the promising performance it offers when consuming only a few
labeled data~\cite{bansal-etal-2020-self,gao2021making}.
However, there are still  significant gaps between few-shot and fully-supervised PLM fine-tuning in many    tasks.
Besides, the performance of few-shot PLM fine-tuning can be sensitive to different sets of training data~\cite{bragg2021flex}.
Therefore, there is a crucial need for 
approaches that make
PLM fine-tuning more label-efficient and robust to selection of training data, especially for applications where labeled data are scarce and expensive to obtain.



Towards this goal, researchers have recently resorted to \emph{active fine-tuning} of PLMs and achieved comparable
performance to fully-supervised methods with much less annotated samples~\citep{ein-etal-2020-active,balm,margatina2021active,yuan-etal-2020-cold}. 
Nevertheless, they usually neglect unlabeled
data, which can be useful for
improving label efficiency for PLM fine-tuning~\cite{du-etal-2021-self}. 
To incorporate unlabeled data into active learning, efforts have been made in the semi-supervised active learning literature~\cite{wang2016cost,rottmann2018deep,simeoni2021rethinking}. However, the query strategies  proposed in these works can return highly redundant samples due to limited representation power, resulting in suboptimal label efficiency.
Moreover, they usually rely on pseudo-labeling to utilize unlabeled data, which requires greater (yet often absent) care to denoise the pseudo labels, otherwise the errors could accumulate and hurt model performance. This issue can be even more severe for PLMs, as the fine-tuning process is often sensitive to different weight initialization and data orderings~\cite{dodge2020fine}. 
Thus, it still remains open and challenging to design robust and label efficient method for active PLM fine-tuning. 

To tackle the above challenges, we propose {\ours}, a new method that improves
the label efficiency and robustness of active PLM fine-tuning. Based on the estimated model  uncertainty, {\ours} tightly couples
\emph{active learning} with \emph{self-training} in each learning round: (1) when the average
uncertainty of a region is low, we trust the model’s predictions and select
its most certain predictions within the region for self-training; (2) when the
average uncertainty of a region is high, indicating inadequate observations
for parameter learning, we actively annotate its most uncertain samples within the
region to improve model performance. Different from existing AL methods that only
leverage uncertainty for querying labels, our uncertainty-driven self-training
paradigm gradually leverages unlabeled data with low uncertainty via self-training,
while reducing the chance of error propagation triggered by highly-uncertain
mis-labeled data.

To further boost model performance for \ours, we design two techniques to improve the
query strategy and suppress label noise, namely
region-aware sampling (RS) and momentum-based memory bank (MMB).
Inspired by the fact that existing uncertainty-based AL methods often end up with 
choosing uncertain yet repetitive
data~\cite{ein-etal-2020-active,margatina2021active}, we design the  region-aware
sampling technique to promote both diversity and
representativeness by leveraging the representation power of PLMs.
Specifically, we first estimate the uncertainties of the unlabeled data with
PLMs, then cluster the data using their PLM representations and  weigh the data by the corresponding uncertainty.
Such a clustering scheme partitions  the embedding space into small sub-regions with an emphasis on highly-uncertain samples.
Finally, by sampling over multiple high-uncertainty regions, our strategy selects data with high uncertainty and low redundancy.

To rectify the erroneous pseudo labels derived by self-training, we design a simple but effective way to select low-uncertainty data for self-training.
Our method is motivated by the fact that fine-tuning  PLMs suffer from  instability issues --- different  initializations and data orders can lead to large variance in  model  performance~\cite{dodge2020fine,zhang2020revisiting,mosbach2021on}. 
However, previous approaches only select pseudo-labeled data based on the prediction of the current round and are thus less reliable. 
In contrast, we maintain a dynamic memory bank to save the predictions of unlabeled samples for later use.
We propose a momentum updating method to dynamically aggregate the predictions from preceding rounds~\cite{laine2016temporal} and select low-uncertainty samples based on aggregated prediction. 
As a result, only the samples with high prediction confidence over multiple rounds will be used for self-training, which mitigates the issue of label noise.
We highlight that our active self-training approach is an efficient substitution to existing AL methods, requiring little extra computational cost.

Our key contributions are: (1) an active self-training paradigm {\ours} that
integrates self-training and active learning to minimize the labeling cost for
fine-tuning PLMs; (2) a region-aware querying strategy to enforce both the
informativeness and the diversity of queried samples during AL; (3) a simple
and effective momentum-based method to leverage the predictions in preceding
rounds to alleviate the label noise in self-training and (4) experiments on 6
benchmarks demonstrating {\ours} improves the label efficiency over existing
self-training and active learning baselines by 56.2\%.

\vspace{-2mm}
\section{Uncertainty-aware Active Self-training}
\vspace{-2mm}

\subsection{Problem Formulation}
\vspace{-1.5mm}
We study active fine-tuning of pre-trained language models for text classification, formulated as follows: Given a small number of labeled samples $\mathcal{X}_l = \{(x_i,y_i)\}_{i=1}^L$ and unlabeled samples $\mathcal{X}_u = \{x_j\}_{j=1}^U$ $(|\mathcal{X}_l| \ll |\mathcal{X}_u|)$, 
we aim to fine-tune a pre-trained language model $f(\bm{x};\theta): \mathcal{X} \rightarrow \mathcal{Y}$ in an interactive way: 
we perform active self-training for $T$ rounds with the total labeling budget $b$. 
In each round, we aim to query  $B=b/T$ samples denoted as $\cB$ from $\cX_u$ to
fine-tune a pre-trained language model $f(\bm{x};\theta)$ with both $\mathcal{X}_l, \cB$ and $\mathcal{X}_u$ to maximize the performance on downstream text classification tasks. Here $\mathcal{X}=\mathcal{X}_l \cup \mathcal{X}_u$ denotes all samples, and $\mathcal{Y}=\{1,2,\cdots,C\}$ is the label set where $C$ is the number of classes.

\subsection{Overview of {\ours} Framework}
\vspace{-1.5mm}
We now present our active self-training paradigm {\ours} underpinned by
estimated uncertainty. 
We begin the active self-training loop by fine-tuning a
BERT $f(\theta^{(\text{0})})$ on the initial labeled data $\cX_L$. Formally, we solve the following optimization problem
\begin{equation}
 \setlength{\abovedisplayskip}{1pt}
\setlength{\belowdisplayskip}{1pt}
\min_{\theta}~\frac{1}{|\cX_L|}\sum_{(\bm{x}_{i}, y_{i}) \in \cX_L} \ell_{\text{CE}}\left(f(\bm{x}_{i}; \theta^{(\text{0})}), y_{i}\right).
\label{eq:init}
\end{equation}
In round $t$ $(1\le t \le T)$ of active self-training, we first calculate the uncertainty score based on a given function $a_i^{(t)} = a(\bm{x}_i, \theta^{(t)})$\footnote{Note that {\ours} is agnostic to the way uncertainty score $a_i^{(t)}$ is computed.} for all $\bm{x}_i \in \cX_u$. Then, we query labeled samples and generate pseudo-labels for unlabeled data $\cX_u$ simultaneously to facilitate self-training.
For each sample $\bx_i$, the pseudo-label $\tilde{y}$ is calculated  based on
the current model's output:
\begin{equation}
 \setlength{\abovedisplayskip}{0.5pt}
\setlength{\belowdisplayskip}{0.5pt}
\tilde{y}=\underset{j \in \mathcal{Y}}{\operatorname{argmax}} \left[{f}(\bx; \theta^{(t)})\right]_{j},
\label{eq:pseudo}
\end{equation}
where $f(\bm{x};\theta^{(t)}) \in \mathbb{R}^C$ is a probability simplex and $[f(\bm{x};\theta^{(t)})]_j$ is the $j$-th entry.
The procedure of {\ours} is summarized in Algorithm \ref{alg:main}.
\begin{algorithm}[t]
	\begin{small}
	\KwIn{Initial labeled samples $\mathcal{X}_l$; Unlabeled samples $\mathcal{X}_u$; Pre-trained LM $f(\cdot; \theta)$, number of active self-training rounds $T$.}
	// \textit{Fine-tune the LM with initial labeled data.} \\
    1. Calculate $\theta^{(0)}$ based on Eq.~\eqref{eq:init}. \\
    2. Initialize the memory bank $g(\bm{x}; \theta^{t})$ based on the current prediction. \\
	// \textit{Conduct active self-training with all data.} \\
	\For{$t = 1, 2, \cdots, T$}{
		{
			1. Run weighted K-Means (Eq.~\eqref{eq:weightedkm1},~\eqref{eq:weightedkm2}) until convergence. \\
			2. Select sample set $\cQ^{(t)}$ for AL and $\cS^{(t)}$ for self-training from $\cX_u$ based on Eq.~\eqref{eq:momprob} or \eqref{eq:momval}. \\
			3. Augment the labeled set $\cX_L = \cX_L \cup \cQ^{(t)}$. \\
            4. Obtain $\theta^{(t)}$ by finetuning $f(\cdot; \theta^{t})$ with $\cL_{\text{ST}}$ ( Eq.~\eqref{eq:st}) using AdamW. \\
            5. Update memory bank $g(\bm{x}; \theta^{t})$ with Eq.~\eqref{eq:probupdate} or \eqref{eq:valupdate}.
		}
	}
	\KwOut{The final fine-tuned model $f(\cdot; \theta^{T})$.}
	\end{small}
	\caption{\begin{small}Training Procedures of \ours. \end{small}}
	\label{alg:main}
\end{algorithm}
\noindent \subsection{Region-aware Sampling for Active Learning on High-uncertainty Data}  
After obtaining the uncertainty for unlabeled data, we aim to query annotation for high-uncertainty samples. 
However, directly sampling the most uncertain samples gives suboptimal results as it tends to query repetitive
data~\citep{ein-etal-2020-active} that  represent
the overall data distribution poorly. 

To tackle this issue, we propose region-aware sampling to capture both
\emph{uncertainty} and \emph{diversity} during active self-training.
Specifically, in the $t$-th round, we first conduct the weighted K-means
clustering~\citep{huang2005automated}, which weights samples based on their
uncertainty. Denote by $K$ the number of clusters and
$\bv_i^{(t)}=\text{BERT}(\bx_i)$ the representation of $\bx_i$ from the
penultimate layer of BERT. The weighted K-means process first initializes the center
of each each cluster $\bm{\mu}_i (1\leq i \leq K)$ via
K-Means++~\cite{kmeans++}. Then, it jointly updates the centroid of each
cluster and assigns each sample to cluster $c_i$ as

\begin{equation}
\setlength{\abovedisplayskip}{0.1pt}
\setlength{\belowdisplayskip}{0.1pt}
 c_i^{(t)} = \underset{k=1, \ldots, K}{\operatorname{argmin}}\left\|\bv_i-\bm{\mu}_{k}\right\|^{2}, 
\label{eq:weightedkm1}
\end{equation}
\vspace{-2mm}
\begin{equation}
\setlength{\abovedisplayskip}{0.1pt}
\setlength{\belowdisplayskip}{0.1pt}
\bm{\mu}_k^{(t)} = \frac{\sum_{\bx_i \in \cC_{k}^{(t)}} a(\bx_i, \theta^{(t)}) \cdot \bv_i^{(t)}}{\sum_{\bx \in \cC_{k}^{(t)}} a(\bx_i, \theta^{(t)}) } 
\vspace{-2mm}
\label{eq:weightedkm2}
\end{equation}

where $\cC_{k}^{(t)} = \{\bx_i^{(t)} | c_i^{(t)} = k \} (k=1,\ldots,K)$ stands for the $k$-th cluster. The above two steps in Eq.~\eqref{eq:weightedkm1},~\eqref{eq:weightedkm2} are repeated until convergence. Compared with vanilla K-Means method, the weighting scheme increases the density of the samples with high uncertainty, thus enabling the K-Means methods to discover clusters with high uncertainty.
After obtaining $K$ regions with the corresponding data $\cC_{k}^{(t)}$, we calculate the uncertainty of each region as

\begin{equation}
 \setlength{\abovedisplayskip}{1pt}
 \setlength{\belowdisplayskip}{1pt}
u_k^{(t)} = U(\cC_{k}^{(t)}) + \beta I(\cC_{k}^{(t)})
\label{eq:region_uncertainty}
\end{equation}
where 
\begin{equation}
\setlength{\abovedisplayskip}{0.5pt}
\setlength{\belowdisplayskip}{0.5pt}
U(\cC_{k}^{(t)}) = \frac{1}{|\cC_{k}^{(t)}|}\sum_{\bx_i\in \cC_{k}^{(t)}}  a(\bx_i, \theta^{(t)}),
\label{eq:data_uncertainty}
\end{equation}
is the average uncertainty of samples and
\begin{equation}
\setlength{\abovedisplayskip}{0.5pt}
\setlength{\belowdisplayskip}{0.5pt}
I(\cC_{k}^{(t)})=-\sum_{j\in C} f_j^{(t)} \log f_j^{(t)}
\label{eq:data_uncertainty}
\end{equation}
is the inter-class diversity within cluster $k$ and $f_{j}^{(t)} = \frac{\sum_i{\mathbbm{1}\{\tilde{y}_i=j}\}}{|\cC_{k}^{(t)}|}$  is the frequency of class $j$ on cluster $k$.
Notably, the term $U(\cC_{k}^{(t)})$ assigns higher score for clusters with more uncertain samples, and $I(\cC_{k}^{(t)})$ grants higher scores for clusters containing samples with more diverse predicted classes  from pseudo labels since such clusters would be closer to the decision boundary.

Then, we rank the clusters in an ascending order in $u_k^{(t)}$.
A high score indicates high uncertainty of the model in these regions, and we need to
actively annotate the member instances to reduce uncertainty and improve the model's performance. 
We adopt a hierarchical sampling strategy: we first
select the $M$ clusters with the highest uncertainty, and then  sample $b' = \lfloor \frac{B}{M} \rfloor$ data with the highest uncertainty to  form the batch $\cQ^{(t)}$.\footnote{If the number of samples in the  $i$-th cluster $\cC_i$ is smaller than $b'$, then we sample all the data within $\cC_i$ and increase the budget for the $(i+1)$-th cluster by $b'-|\cC_i|$. }
\begin{equation}
\begin{small}
\begin{aligned}
 \setlength{\abovedisplayskip}{1pt}
\setlength{\belowdisplayskip}{1pt}
\cK_a^{(t)} & = \underset{k \in \{1, \ldots, K\}}{\operatorname{top-M}} \ u_k^{(t)}, \\
\cQ^{(t)} &= \bigcup\limits_{k \in \cK_{a}^{(t)}}\cC_{a,k}^{(t)}
\ \operatorname{where} \ \cC_{a,k}^{(t)} = \underset{\bm{x}_{i} \in \cC_{k}^{(t)}}{\operatorname{Top-b}'} \ a(\bm{x}_i, \theta^{(t)}).
\end{aligned}
\label{eq:stage1}
\end{small}
\vspace{-1mm}
\end{equation}
We remark that such a hierarchical sampling strategy queries most uncertain samples from \emph{different} regions, thus the uncertainty and diversity of queried samples can be both achieved.

\noindent \subsection{Self-training over Confident Samples from Low-uncertainty Regions}  
\vspace{-1mm}
For self-training, we aim to select unlabeled samples which are \emph{most likely} to have been correctly classified by the current model. This requires the sample to have low  uncertainty. 
Therefore, we select the top $k$ samples from the $M$ lowest uncertainty regions to form the acquired batch $\cS^{(t)}$:

\vspace{-0.2cm}

\begin{small}
\begin{equation}
\begin{aligned}
 \setlength{\abovedisplayskip}{1pt}
\setlength{\belowdisplayskip}{1pt}
\cC_s^{(t)} &= \bigcup\limits_{k \in \cK_s^{(t)}} \cC_{k}^{(t)} \ \operatorname{where} \ \cK_s^{(t)} = \underset{k \in \{1, \ldots, K\}}{\operatorname{bottom-M}} \ u_k^{(t)}, \\
\cS^{(t)} &= \underset{\bm{x}_{i} \in \cC_s^{(t)}}{\operatorname{bottom-k}} \ a(\bm{x}_i, \theta^{(t)}).
\end{aligned}
\label{eq:st_set}
\end{equation}
\end{small}

\vspace{-0.44cm}

\noindent \textbf{Momentum-based Memory Bank for Self-training.} 
As PLMs are sensitive to the stochasticity involved in
fine-tuning, the model suffers from the instability issue --- different weight
initialization and data orders may result in different predictions on the same dataset~\cite{dodge2020fine}. 
Additionally, if the model gives inconsistent predictions in different rounds
for a specific sample, then it is potentially uncertain about the sample, and
adding it to the training set may harm the active self-training process.
For example, for a two-class classification problem, suppose we obtain $f(\bm{x};\theta^{(t-1)})=[0.65, 0.35]$ for sample $\bx$ the round $(t-1)$ and $f(\bm{x};\theta^{(t)})=[0.05, 0.95]$ for the round $t$. 
Although the model is quite `confident' on the class of $\bm{x}$ when we only consider the result of the round $t$, it gives contradictory predictions over these two consecutive rounds, which indicates that the model is actually uncertain to which class $\bm{x}$ belongs.

To effectively mitigate the noise and stabilize the active self-training
process, we maintain a dynamic memory bank to save the results from previous
rounds, and use momentum update~\cite{He_2020_CVPR,laine2016temporal} to
aggregate the results from both the previous and current rounds. 
Then, during active self-training, we will select samples with the highest aggregated score. In this way, only those samples that the model is certain about over all \emph{previous rounds} will be selected for self-training. 
We design two variants for the memory bank, namely \emph{prediction-based} and
\emph{value-based}  aggregation.

\noindent  \textbf{Prediction based Momentum Update.} We adopt an exponential moving average approach to aggregate the prediction $g(\bm{x};\theta^{(t)})$ on round $t$ as 
\begin{equation}
 \setlength{\abovedisplayskip}{1pt}
\setlength{\belowdisplayskip}{1pt}
g(\bm{x};\theta^{(t)}) = m_t \times f(\bm{x};\theta^{(t)}) + (1-m_t) \times g(\bm{x};\theta^{(t-1)}),
\label{eq:probupdate}
\end{equation}
where $m_t = (1 - \frac{t}{T})m_L + \frac{t}{T}m_H \ (0 < m_L \leq m_H \leq1)$ is a momentum coefficient. We gradually increase the weight for models on later rounds,  
since they are trained with more labeled data thus being able to provide more reliable predictions.
Then, we calculate the uncertainty based on $g(\bm{x};\theta^{(t)})$ and
rewrite Eq. \eqref{eq:st_set} and \eqref{eq:pseudo} as 
\begin{equation}
\begin{aligned}
\setlength{\abovedisplayskip}{0.1pt}
\setlength{\belowdisplayskip}{0.1pt}
\cS^{(t)}&=\underset{\bm{x}_{i} \in \cC_s^{(t)}  }{\operatorname{bottom-k }} \ a\left(\bm{x}_i, g(\bm{x};\theta^{(t)}), \theta^{(t)}\right) \\
\tilde{y}&=\underset{j \in \mathcal{Y}}{\operatorname{argmax}} \left[{g}(\bx; \theta^{(t)})\right]_{j},
\end{aligned}
\label{eq:momprob}
\end{equation}

\noindent  \textbf{Value-based Momentum Update.} For methods that do not directly use prediction for uncertainty estimation, we aggregate the uncertainty value as 
\begin{equation}
 \setlength{\abovedisplayskip}{0.5pt}
\setlength{\belowdisplayskip}{0.5pt}
\begin{small}
g(\bm{x};\theta^{(t)}) = m_t \times a(\bm{x};\theta^{(t)}) + (1-m_t) \times g(\bm{x};\theta^{(t-1)}).
\end{small}
\label{eq:valupdate}
\end{equation}
Then, we use Eq.~\eqref{eq:valupdate} to sample low-uncertainty data for self-training as\footnote{Other choices such as soft pseudo label~\citep{xie2020unsupervised,liang2020bond} is also applicable.}
\begin{equation}
 \setlength{\abovedisplayskip}{0.5pt}
\setlength{\belowdisplayskip}{0.5pt}
\begin{aligned}
\cS^{(t)}&=\underset{\bm{x}_{i} \in \cC_s^{(t)}}{\operatorname{bottom-k }} \ g(\bm{x}_i, \theta^{(t)}), \\
\tilde{y}&=\underset{j \in \mathcal{Y}}{\operatorname{argmax}} \left[f(\bx; \theta^{(t)})\right]_{j}.
\end{aligned}
\label{eq:momval}
\end{equation}

By aggregating the prediction results over previous rounds, we filter the sample with inconsistent predictions to suppress noisy labels. 
\vspace{-1mm}
\subsection{Model Learning and Update}
\vspace{-1mm}
After obtaining both the labeled data and pseudo-labeled data, we fine-tune a new pre-trained BERT model $\theta^{(t+1)}$ on them. 
Although we only include low-uncertainty samples during self-training, it is
difficult to eliminate all the wrong pseudo-labels, and such mislabeled
samples can still hurt model performance.
To suppress such label noise, we use a threshold-based strategy to further remove noisy labels by selecting samples that agree with the corresponding pseudo labels. The loss objective of optimizing $\theta^{(t+1)}$ is

\begin{equation}
 \setlength{\abovedisplayskip}{1pt}
\setlength{\belowdisplayskip}{1pt}
\begin{small}
\begin{aligned}
\mathcal{L}_{\text{ST}}=& \frac{1}{\left|\mathcal{X}_{L} \cup \cQ^{(t)}\right|} \sum_{\boldsymbol{x}_{i} \in \mathcal{X}_{L}\cup \cQ^{(t)}} \ell_{\text{CE}}\left({f}(\bx_i; \theta^{(t+1)}), y_{i}\right) \\
&+\frac{\lambda}{\left|\cS^{(t)}\right|} \sum_{\tilde{\boldsymbol{x}}_{i} \in \cS^{(t)}} \omega_i \ell_{\text{CE}}\left({f}(\tilde{\boldsymbol{x}}_{i}; \theta^{(t+1)}), \tilde{y}_{i}\right),
\end{aligned}
\end{small}
\label{eq:st}
\end{equation}
where $\lambda$ is a hyper-parameter balancing the weight between clean  and pseudo labels, and $\omega_i = \mathbbm{1}\{\left[f(\bm{x}_i;  \theta^{(t+1)})\right]_{\tilde{y}_i}>\gamma \}$ stands for the thresholding function.

 



\noindent  \textbf{Complexity Analysis.} The running time of {\ours} is mainly consisted of two parts: the inference time $O(|\cX_u|)$ and the time for K-Means clustering $O(dK|\cX_u|)$, where $d$ is the dimension of the BERT feature $\bv$. 
For self-training, the size of the memory bank $g(\bm{x}; \theta)$ is proportional to $|\cX_u|$, while the extra computation of maintaining this dictionary is \emph{ignorable} since we do not inference over the unlabeled data for multiple times in each round as BALD~\cite{gal2017deep} does. The running time of {\ours} will be shown in section \ref{sec:time}.

\section{Experiments}
\label{sec:exp}
\vspace{-1.5mm}
\subsection{Experiment Setup}
\label{exp_setup}
\vspace{-1mm}
\noindent\textbf{Tasks and Datasets.}
In our main experiments, we use  4 datasets, including \emph{SST-2}~\cite{socher-etal-2013-recursive} for sentiment analysis, \emph{AGNews}~\cite{zhang2015character} for news topic classification,  \emph{Pubmed-RCT}~\cite{dernoncourt-lee-2017-pubmed} for medical abstract classification, and 
\emph{DBPedia}~\cite{zhang2015character} for wikipedia topic classification. 
For weakly-supervised text classification, we choose 2 datasets, namely  \emph{TREC}~\cite{li2002learning} and \emph{Chemprot}~\cite{chemprot} from the WRENCH benchmark~\cite{zhang2021wrench} for evaluation.
The statistics are shown in Table \ref{tab:dataset}.

\noindent\textbf{Active Learning Setups.} Following~\cite{yuan-etal-2020-cold}, we set the number of rounds $T=10$, the overall budget for all datasets $b=1000$ and the initial size of the labeled $|\cX_l|$ is set to 100.
In each AL round, we sample a batch of 100 samples from the unlabeled set $\cX_u$ and query their labels.  
Since large development sets are impractical in low-resource settings~\cite{kann2019towards}, we keep the size of development set as 1000, which is the same as the labeling budget\footnote{This is often neglected in previous low-resource AL studies, and we 
highlight it to ensure the true low-resource setting.}. For weakly-supervised text classification, since the datasets are much smaller, we keep the labeling budget and the size of development set to $b=500$.
\begin{table}[t]
    \centering
\begin{adjustbox}{max width=0.485\textwidth}
\begin{tabular}{c|c|c|c|c|c}
\toprule
    \bf Dataset & \bf Label Type & \bf \# Class & \bf \# Train & \bf \# Dev & \bf \#Test  \\\midrule
    SST-2 & Sentiment  & 2 & 60.6k & 0.8k & 1.8k \\
    AG News & News Topic  & 4 & 119k & 1k & 7.6k \\
    Pubmed & Medical Abstract  & 5 & 180k & 1k & 30.1k \\
    DBPedia & Wikipedia Topic & 14 & 280k & 1k & 70k \\ \midrule
    TREC & Question & 6 & 5.0k & 0.5k & 0.5k \\
    Chemprot & Medical Abstract & 10 & 12.8k & 0.5k & 1.6k \\
    \bottomrule
\end{tabular}
\end{adjustbox}
\vspace{-1ex}
\caption{Dataset Statistics. For DBPedia, we randomly sample 20k sample from each class due to the limited computational resource.}
\vspace{-3ex}
\label{tab:dataset}
\end{table}

\noindent\textbf{Implementation Details.}
We choose RoBERTa-base~\cite{liu2019roberta} from the HuggingFace codebase~\cite{huggingface} as the backbone for {\ours} and all baselines except for Pubmed and Chemprot, where we use SciBERT~\cite{beltagy2019scibert},  a BERT model pre-trained on scientific corpora.
In each round, we train
from scratch to avoid  overfitting the data
collected in earlier rounds as observed by \citet{hu2018active}.
More details are in Appendix \ref{appendix:exp}.

\noindent\textbf{Hyperparameters.}
The hyperparameters setting is in Appendix~\ref{appendix:ours}. 
In the $t$-th round of active self-training, we increase the number of
pseudo-labeled samples by $k$, where $k$ is $500$ for TREC and Chemprot, $3000$ for SST-2 and Pubmed-RCT, and $5000$ for others. For the momentum factor, we tune $m_L$ from $[0.6, 0.7, 0.8]$ and $m_H$ from $[0.8, 0.9, 1.0]$ and report the best $\{m_L, m_H\}$ based on the performance of the development set. 

\noindent\textbf{Baselines.}

\noindent\textbf{Self-training Methods:} (1) \textbf{Self-training (ST, \citet{lee2013pseudo})}: It is the vanilla self-training method that generates pseudo labels for unlabeled data. (2) \textbf{UST}~\cite{mukherjee2020uncertainty,rizve2021in}: It is an uncertainty-based self-training method that only uses low-uncertainty data for self-training. 
(3) \textbf{COSINE}~\cite{yu-etal-2021-fine}: It uses self-training to fine-tune LM with weakly-labeled data, which achieves SOTA performance on various text datasets in WRENCH benchmark~\cite{zhang2021wrench}.
Note that for these two baselines, we \emph{randomly sample} $b$ labeled data as the initialization.  

\noindent\textbf{Active Learning Methods:} (1) \textbf{Random}: It acquires annotation randomly, which serves as a baseline for all methods. (2) \textbf{Entropy}~\cite{holub2008entropy}: It is an uncertainty-based method that acquires annotations on samples with the highest predictive entropy. 
(3) \textbf{BALD}~\cite{gal2017deep}: It is also an uncertainty-based method, which calculates \emph{model uncertainty} using MC Dropout~\cite{mcdropout}.
(4) \textbf{BADGE}~\cite{badge}: It first selects high uncertainty samples then uses KMeans++ over the gradient embedding to sample data.
(5) \textbf{ALPS}~\cite{yuan-etal-2020-cold}: It uses the masked language model (MLM) loss of BERT to query labels for samples.
(6) \textbf{CAL}~\cite{margatina2021active} is the most recent AL method for pre-trained LMs. It calculates the uncertainty of each sample based on the KL divergence between the prediction of itself and its neighbors' prediction. \\
\noindent\textbf{Semi-supervised Active Learning (SSAL) Methods:}
(1) \textbf{ASST}~\cite{tomanek2009semi,simeoni2021rethinking} is an active semi-supervised learning method that jointly queries labels for AL and samples pseudo labels for self-training. 
(2) \textbf{CEAL}~\cite{wang2016cost} acquires annotations on informative samples, and uses high-confidence samples with predicted pseudo labels for weights updating.
(3) \textbf{BASS}~\cite{rottmann2018deep} is similar to CEAL, but use MC dropout for querying labeled sample.
(4) \textbf{REVIVAL}~\cite{Guo_2021_ICCV} is the most recent SSAL method, which uses an adversarial loss to query samples and leverage label propagation to exploit adversarial examples.

\noindent \textbf{Our Method}: We experiment with both Entropy and CAL as
uncertainty measures for  {\ours}. Note that when compared with active learning baselines, we do not augment the train set with pseudo-labeled data (Eq.~\eqref{eq:st_set}) to ensure fair comparisons.


%

\begin{figure*}[t]
    \vspace{-2mm}
        \centering
        \subfigure[SST-2, AL]{
    \includegraphics[width=0.248\textwidth]{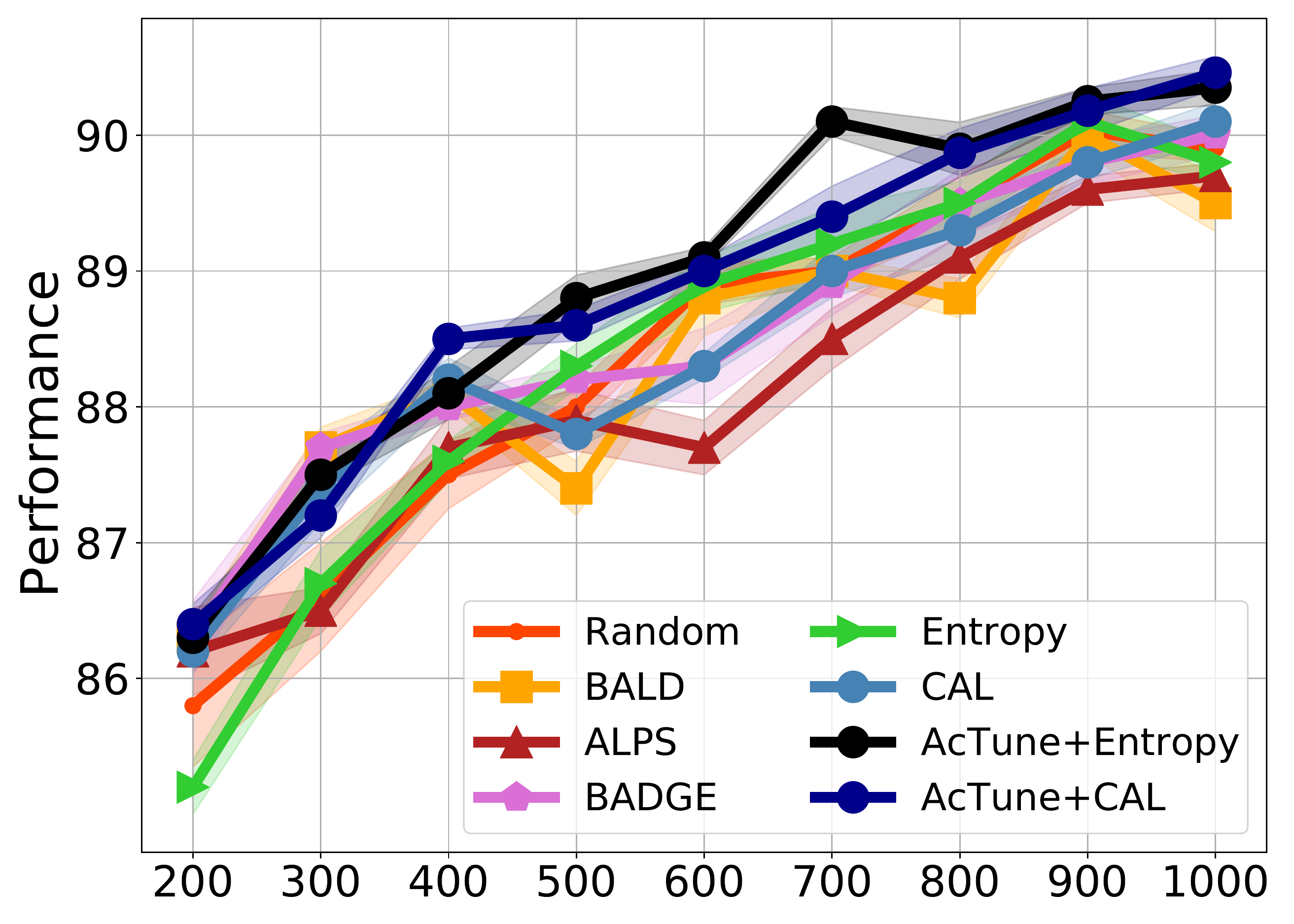}
            \label{fig:sst_al}
        }\hspace{-3.5mm}
        \subfigure[AG News, AL]{
            \includegraphics[width=0.248\textwidth]{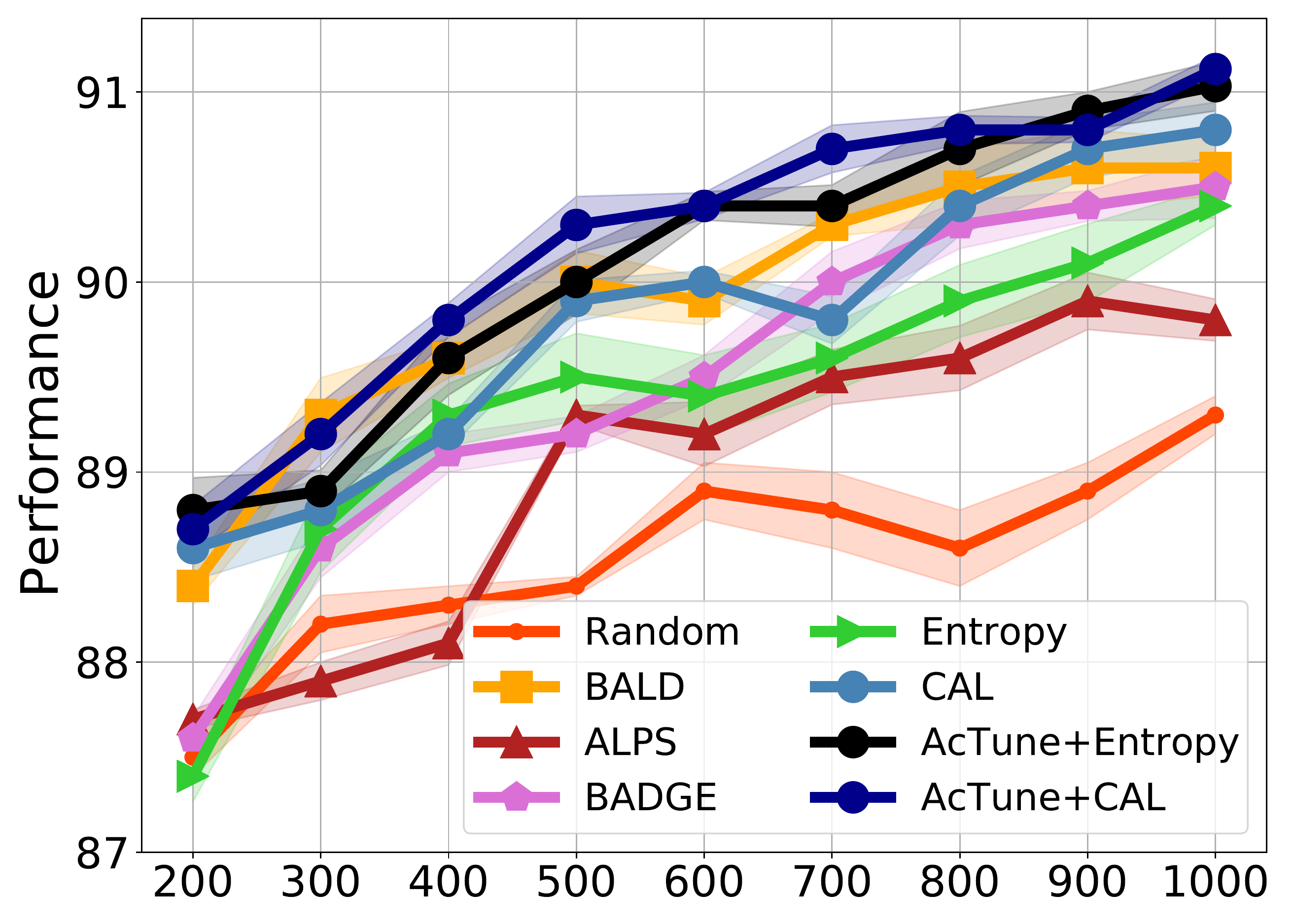}
            \label{fig:ag_al}
        }\hspace{-3.5mm}
        \subfigure[Pubmed, AL]{
            \includegraphics[width=0.248\textwidth]{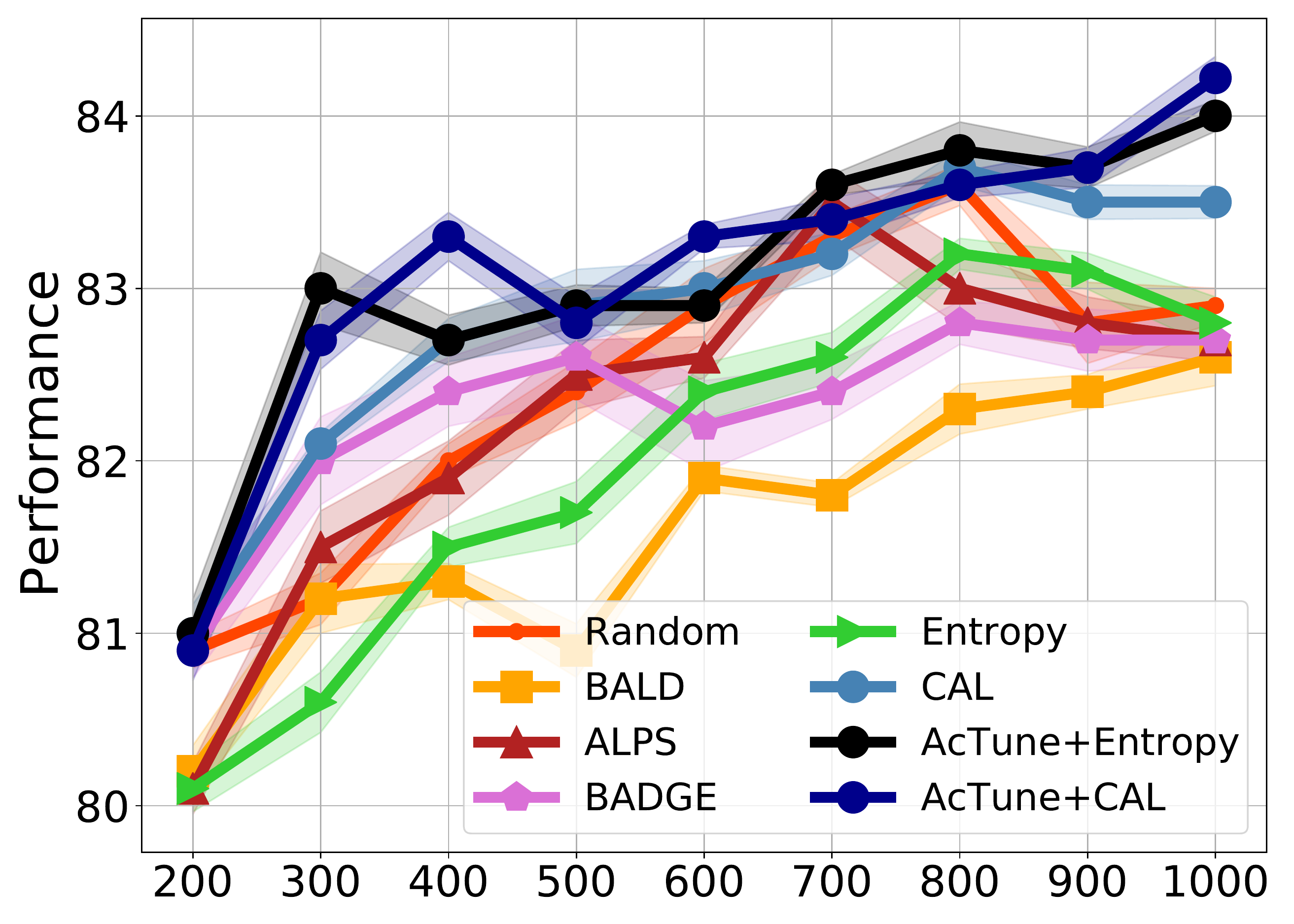}
            \label{fig:pubmed_al}
        }\hspace{-3.5mm}
            \vspace{-2ex}
         \subfigure[DBPedia, AL]{
    \includegraphics[width=0.248\textwidth]{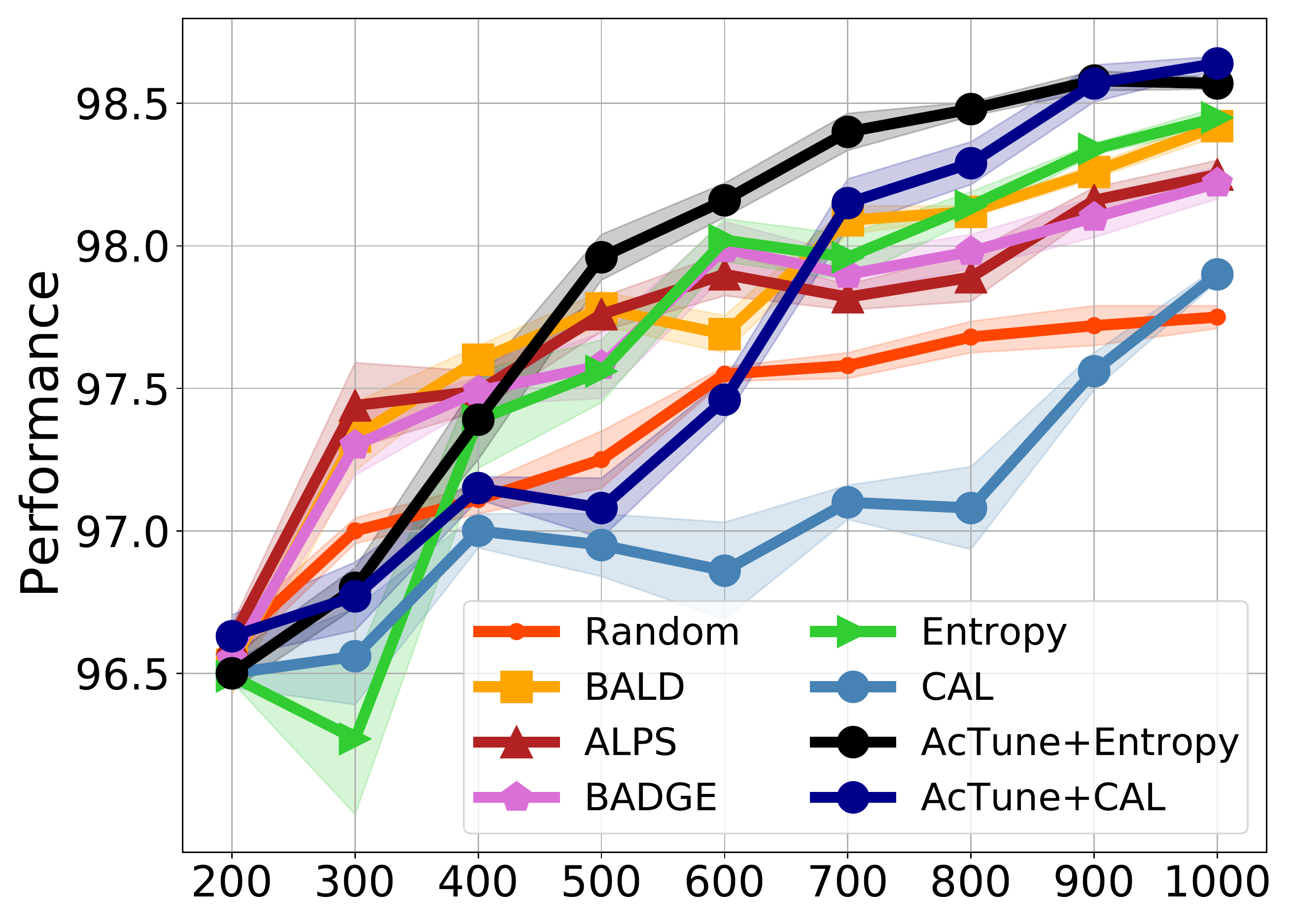}
            \label{fig:dbpedia_al}
        }
        \subfigure[SST-2, SSAL]{
            \includegraphics[width=0.248\textwidth]{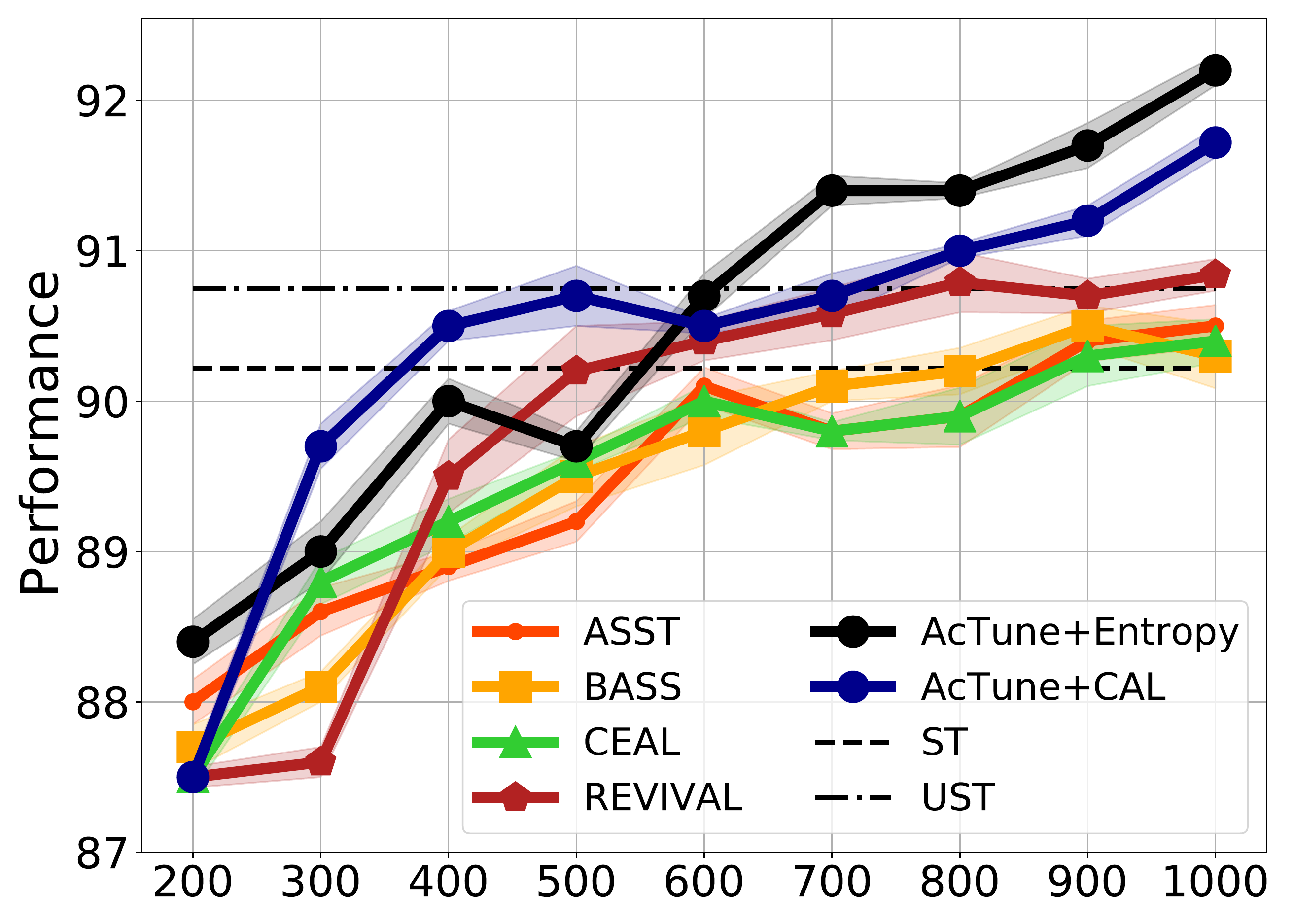}
            \label{fig:sst_ssal}
        }\hspace{-3.5mm}
        \subfigure[AG News, SSAL]{
            \includegraphics[width=0.248\textwidth]{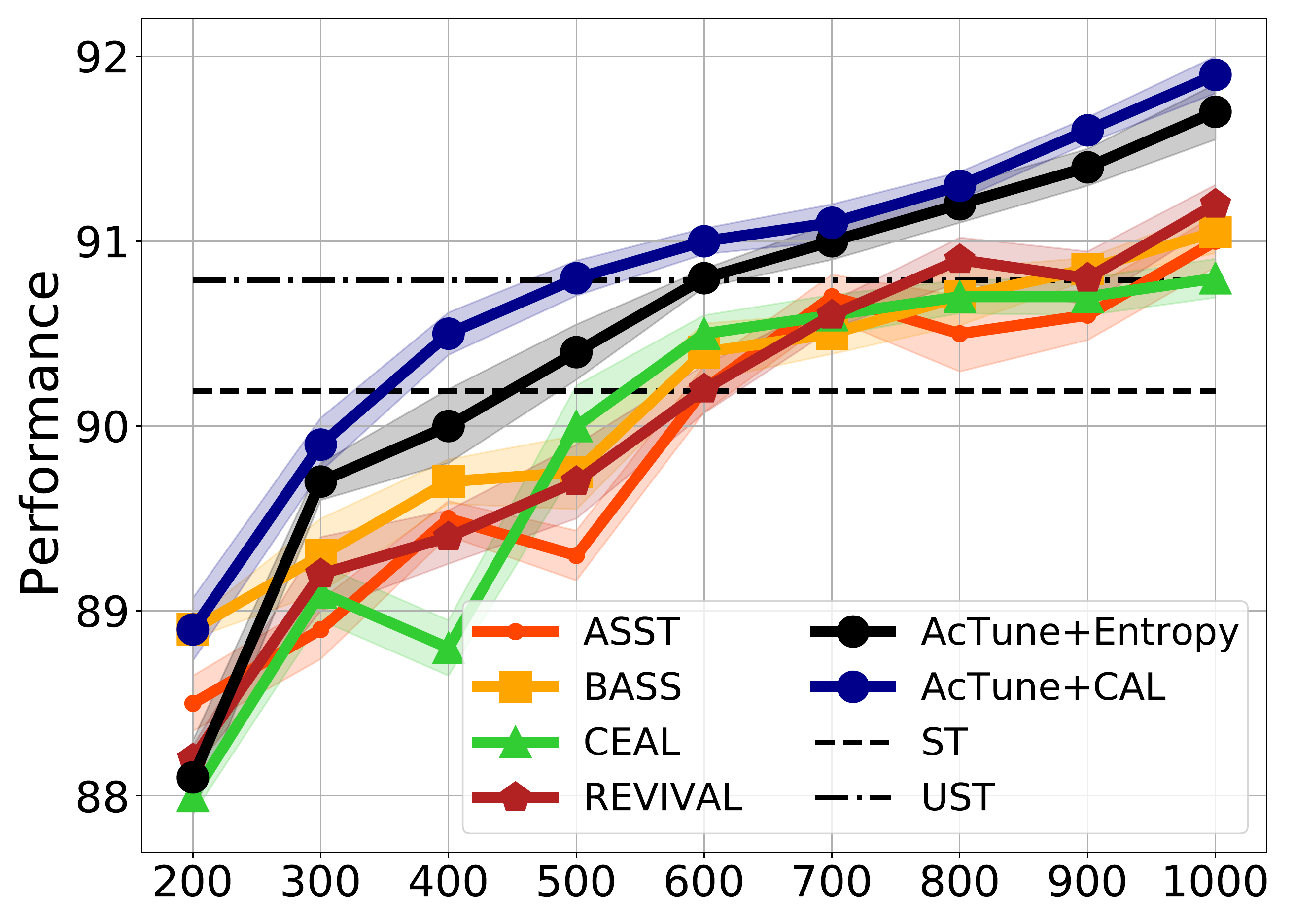}
            \label{fig:ag_ssal}
        }\hspace{-3.5mm}
         \subfigure[Pubmed, SSAL]{
    \includegraphics[width=0.248\textwidth]{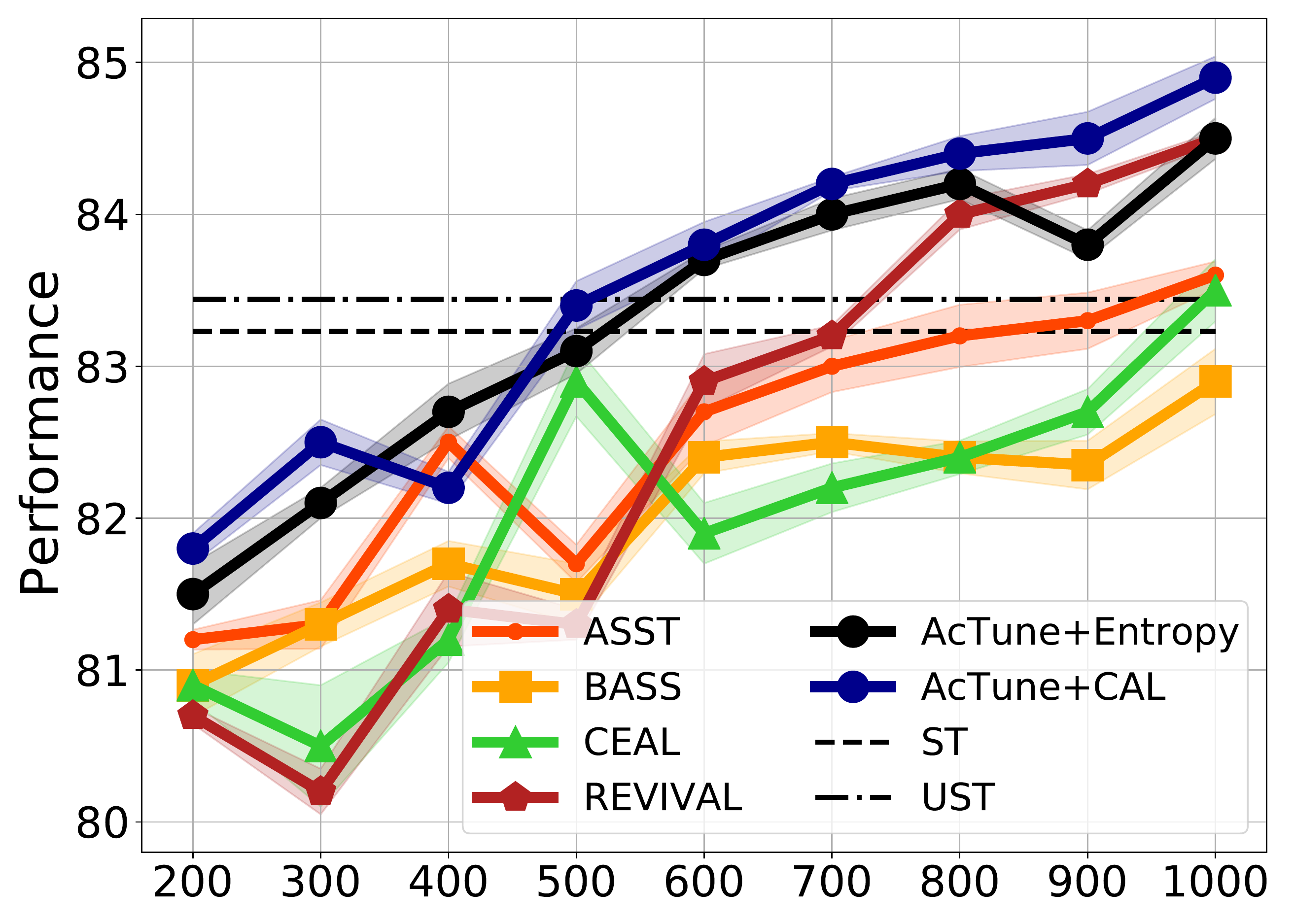}
            \label{fig:pubmed_ssal}
        }\hspace{-3.5mm}
        \subfigure[DBPedia, SSAL$^\dagger$]{
            \includegraphics[width=0.248\textwidth]{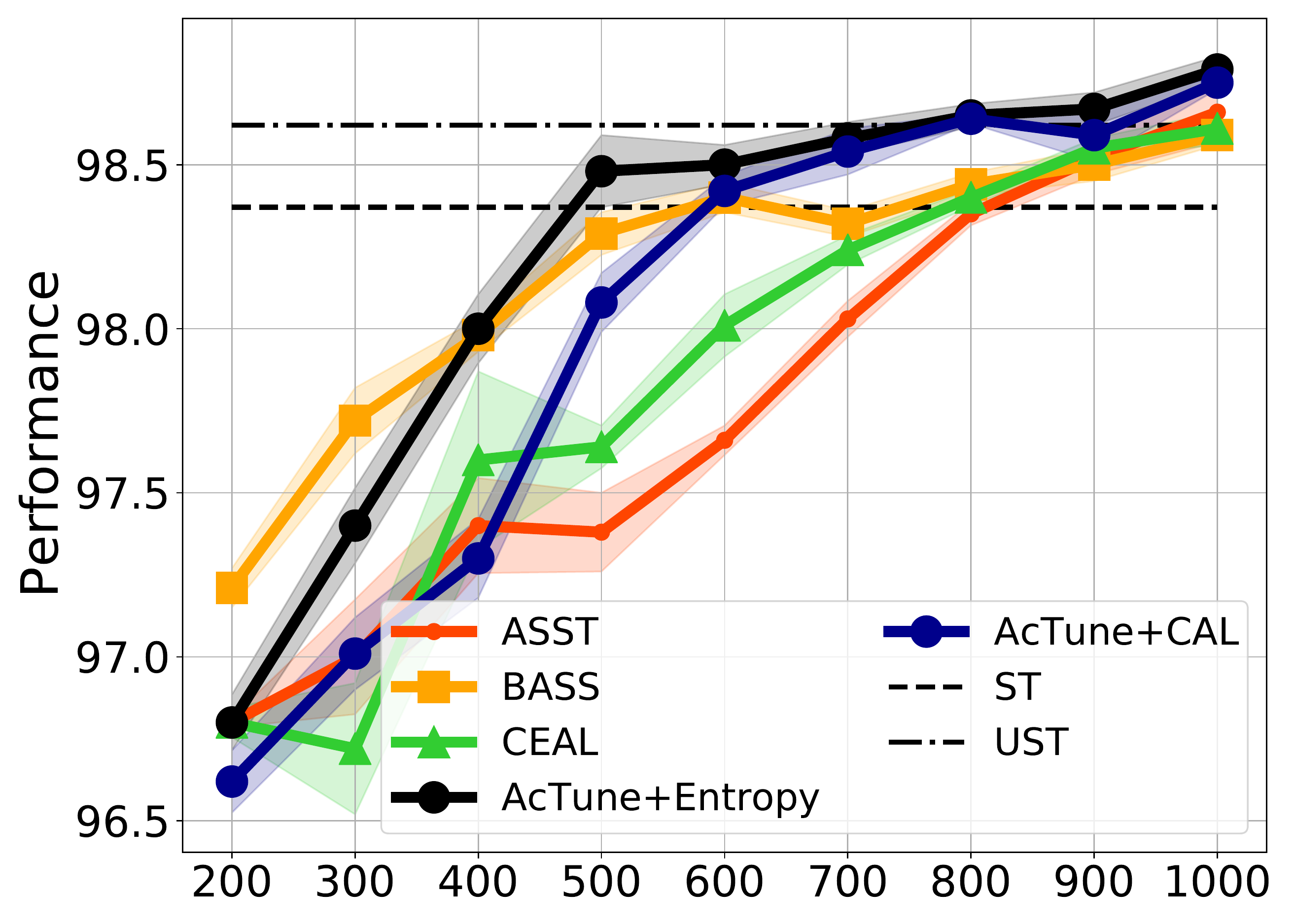}
            \label{fig:dbpedia_ssal}
        }
        \vspace{-2ex}
        \caption{The comparision of {\ours} with active learning,  semi-supervised active learning and self-training baselines. The first row is the result under active learning setting (AL, i.e. no unlabeled data is used), the second row is the result under semi-supervised active learning (SSAL) setting. 
        The metric is accuracy.  $^\dagger$: REVIVAL causes OOM error for DBPedia dataset.}\label{fig:main}
         \vspace{-3ex}
\end{figure*}

\subsection{Main Result}
\label{sec:main}
\begin{figure}[t]
        \centering
    \includegraphics[width=0.4\textwidth]{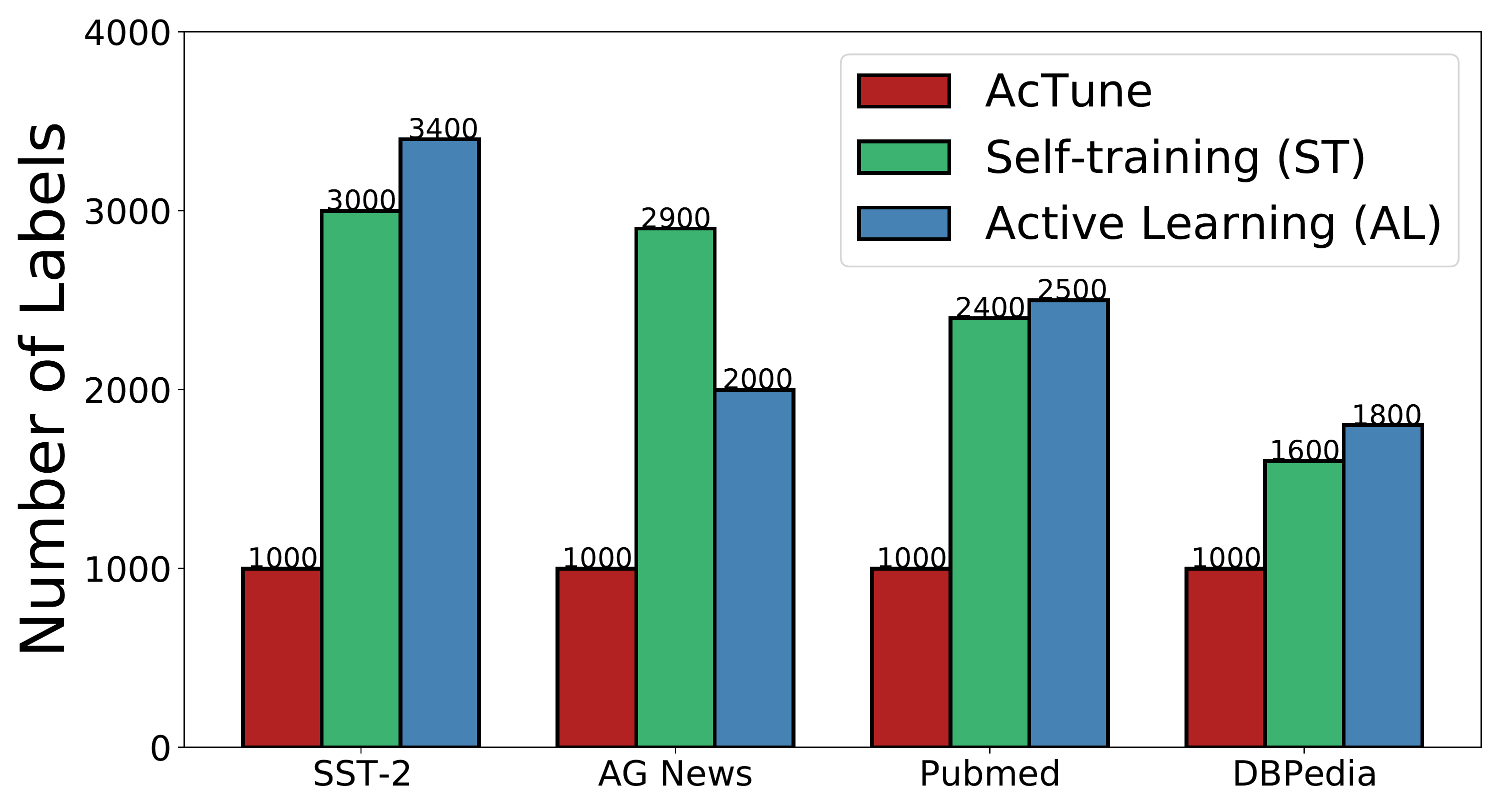}
            \label{fig:entropy}
        \caption{The label-efficiency of {\ours} compared with AL and self-training baselines. According to Fig.~\ref{fig:main}, the best AL method is Entropy for DBPedia and CAL for others. }
        \label{fig:efficiency}
         \vspace{-2ex}
\end{figure}

\vspace{-1mm}
Figure~\ref{fig:main} reports the performance of {\ours} and the
baselines on 4 benchmarks. From the results, we have the
following observations:

\noindent $\bullet$ {\ours} consistently outperforms baselines in most of the cases. 
Different from studies in the computer vision (CV) domain~\cite{simeoni2021rethinking} where the model does not perform well in the low-data regime, pre-trained LM has achieved competitive performance with only a few labeled data, which makes further improvements to the vanilla fine-tuning challenging.  
Nevertheless, {\ours} surpasses baselines in more than 90\% of the rounds and achieves 0.4\%-0.7\% and 0.3\%-1.5\% absolute gain at the end of AL and SSAL respectively.
Figure~\ref{fig:efficiency} quantitatively measures the number of labels needed for the most advanced active learning model and self-training model (UST) to outperform {\ours} with 1000 labels. These baselines need >2000 clean labeled samples to reach the performance as ours.
{\ours} saves on average \textbf{56.2\%}  and \textbf{57.0\%} of the labeled samples than most advanced active learning and self-training baselines respectively, which justifies its promising performance under low-resource scenarios. 
Such improvements show the merits of two key designs under our active self-training framework: the region-aware sampling for active learning and the momentum-based memory bank for robust self-training, which will be discussed in the  section \ref{sec:ablation}.

\noindent $\bullet$ Compared with the previous AL baselines, {\ours} can bring consistent performance gain, while previous semi-supervised active learning methods cannot. For instance, BASS is based on BALD for active learning, but sometimes it performs even worse than BALD with the same number of labeled data (see Fig. \ref{fig:ag_al} and Fig. \ref{fig:ag_ssal}). 
This is mainly because previous methods simply combine noisy pseudo labels with clean labels for training without explicitly rectifying the wrongly-labeled data, which will cause the LM to overfit these hazardous labels.
Moreover, previous methods do not exploit momentum updates to stabilize the learning process, as there are oscillations in the beginning rounds. In contrast, {\ours} achieves a more stable learning process and enables an active self-training process to benefit from more labeled data.

\noindent $\bullet$ The self-training methods (ST \& UST) achieve superior performance with limited labels. However, they mainly focus on leveraging unlabeled data for improving the performance, while our results demonstrate that adaptive selecting the most useful data for fine-tuning is also important for improving the performance. With a powerful querying policy, {\ours} can improve  these self-training baselines by 1.05\% in terms of accuracy on average.




\begin{figure}[ht]
        \centering
        \subfigure[TREC]{
    \includegraphics[width=0.23\textwidth]{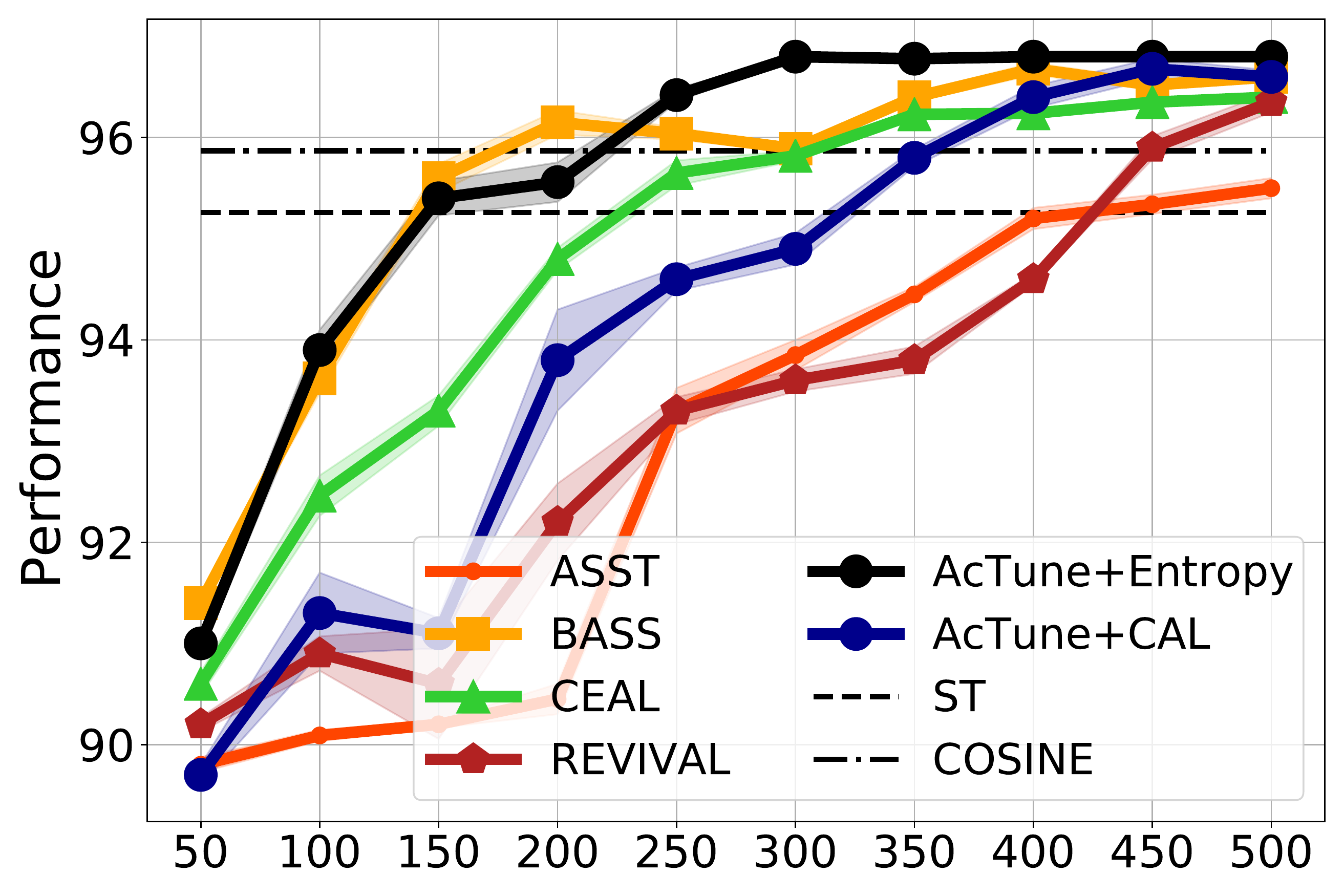}
            \label{fig:trec_al}
        }\hspace{-3mm}
        \subfigure[Chemprot]{
            \includegraphics[width=0.23\textwidth]{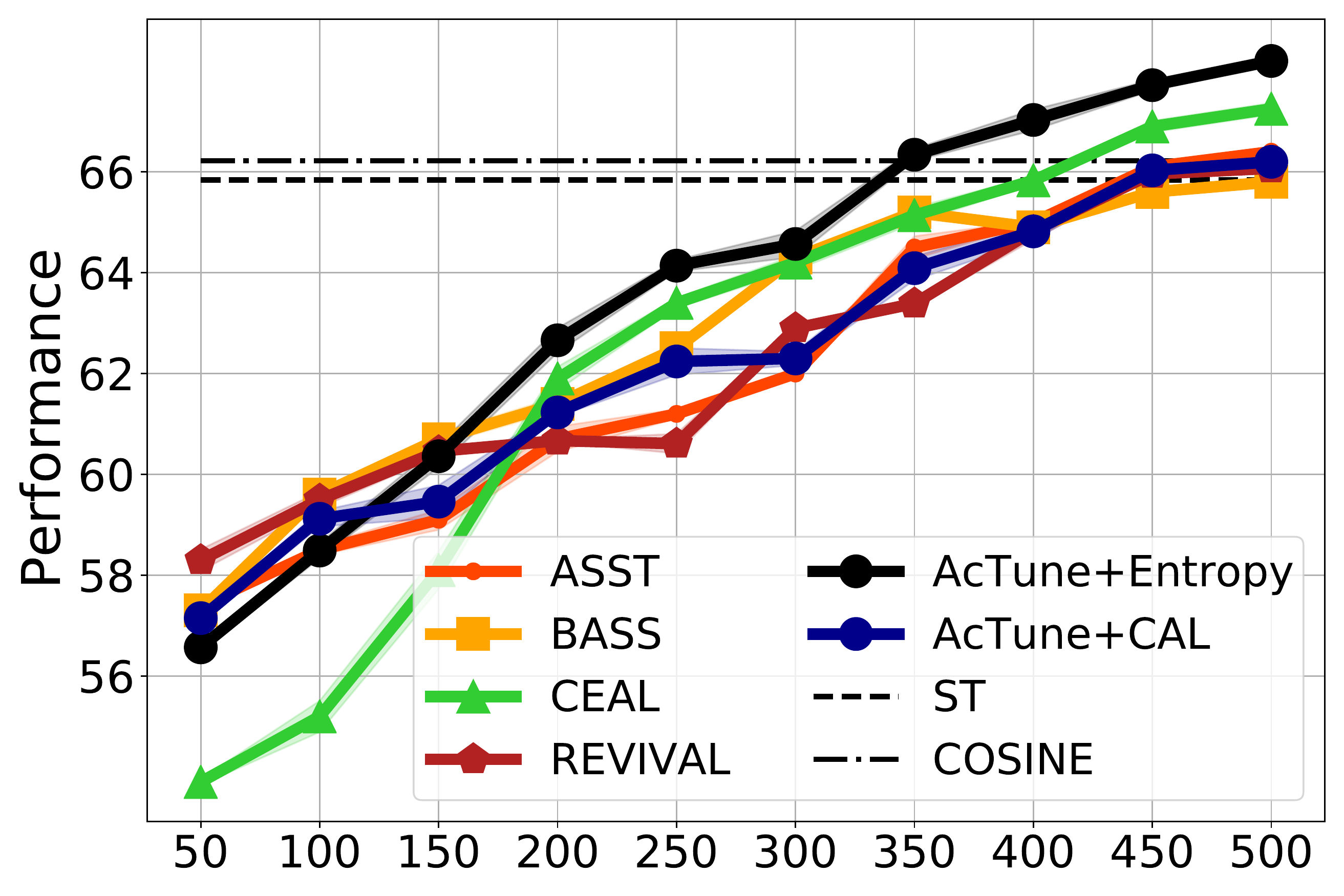}
            \label{fig:chem_al}
        }
        \vspace{-2ex}
        \caption{The comparison of {\ours} and baselines on weakly-supervised classification tasks.} 
        \label{fig:weak}
         \vspace{-3ex}
\end{figure}


\begin{figure*}[t]
    \vspace{-2mm}
        \centering
        \subfigure[Effect of $\beta$]{
            \includegraphics[width=0.19\textwidth]{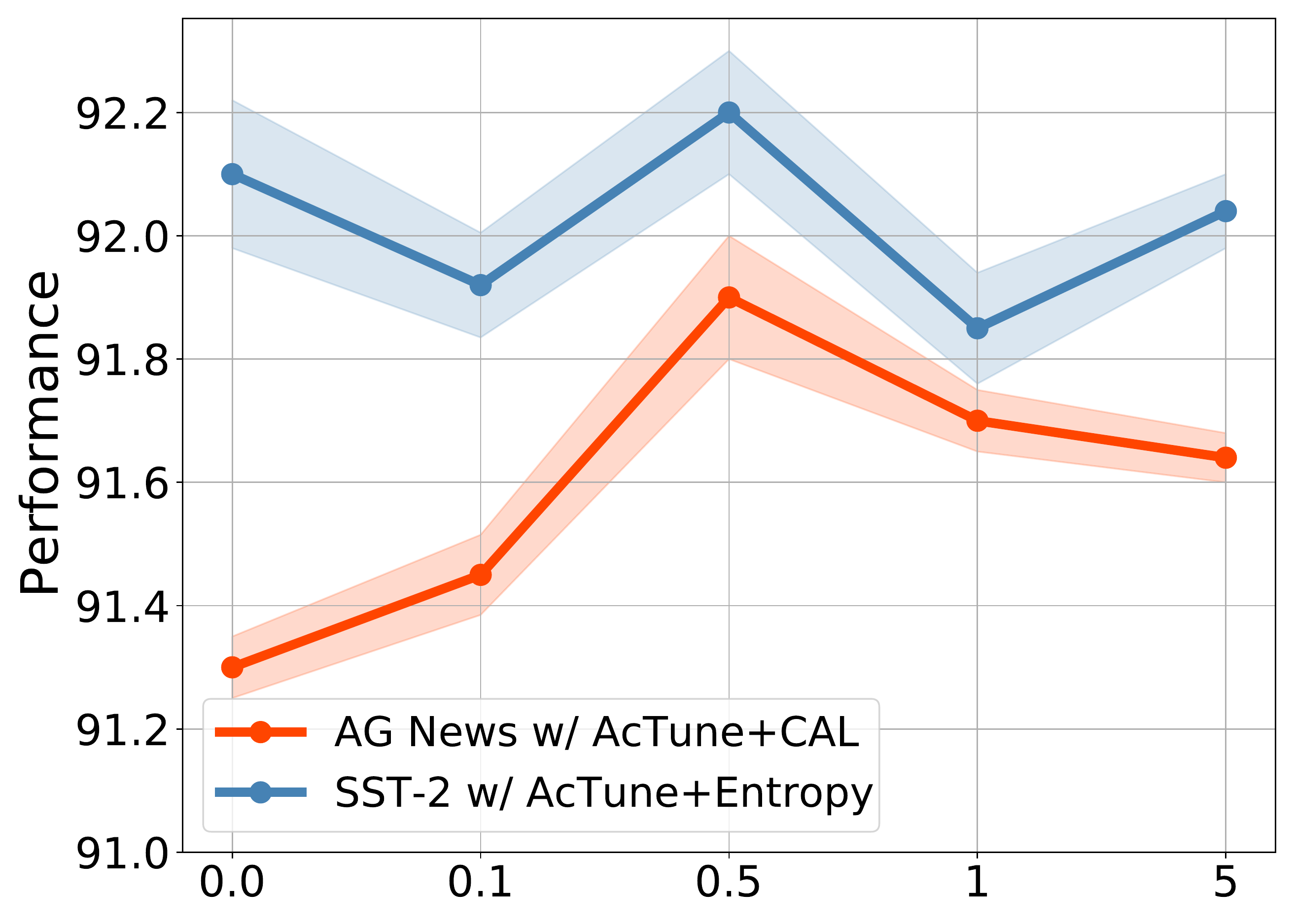}
            \label{fig:param_beta}
        }\hspace{-2.3mm}
        \subfigure[Effect of $K$]{
            \includegraphics[width=0.19\textwidth]{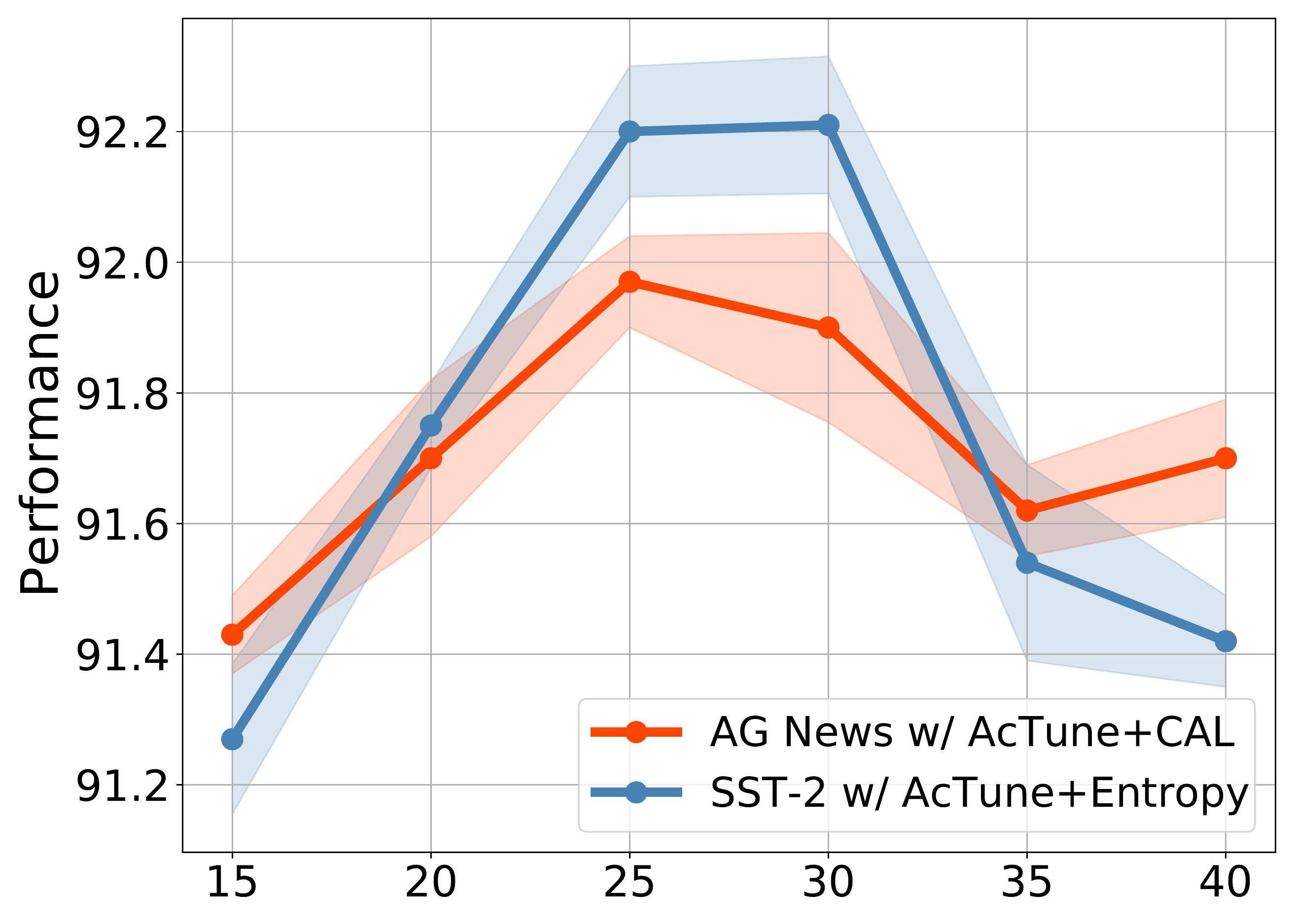}
            \label{fig:param_K}
        }\hspace{-2.3mm}
                 \subfigure[Acc. of PL]{
            \includegraphics[width=0.19\textwidth]{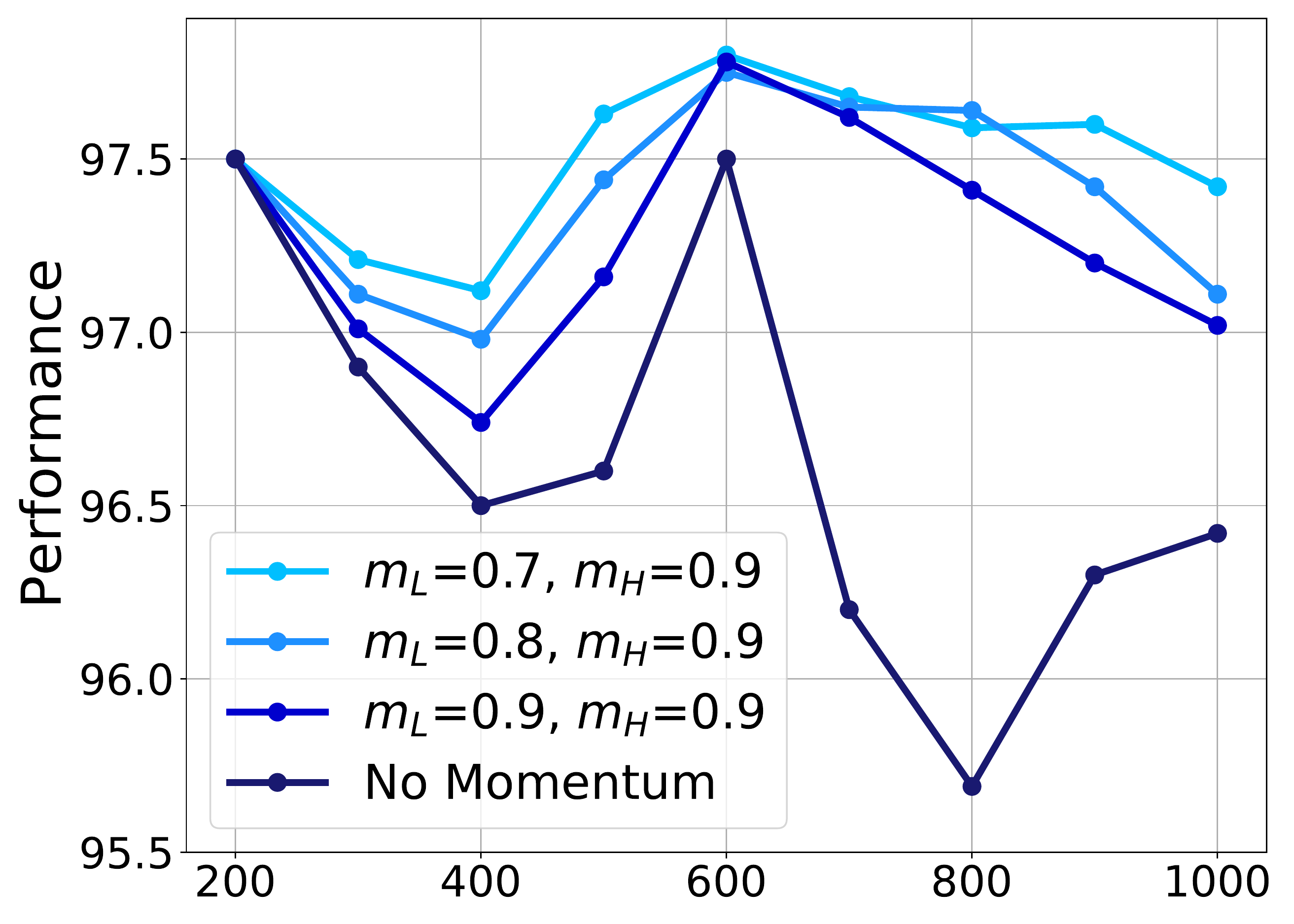}
            \label{fig:acc_pseudo}
        }\hspace{-2.3mm}
        \subfigure[Entropy]{
    \includegraphics[width=0.19\textwidth]{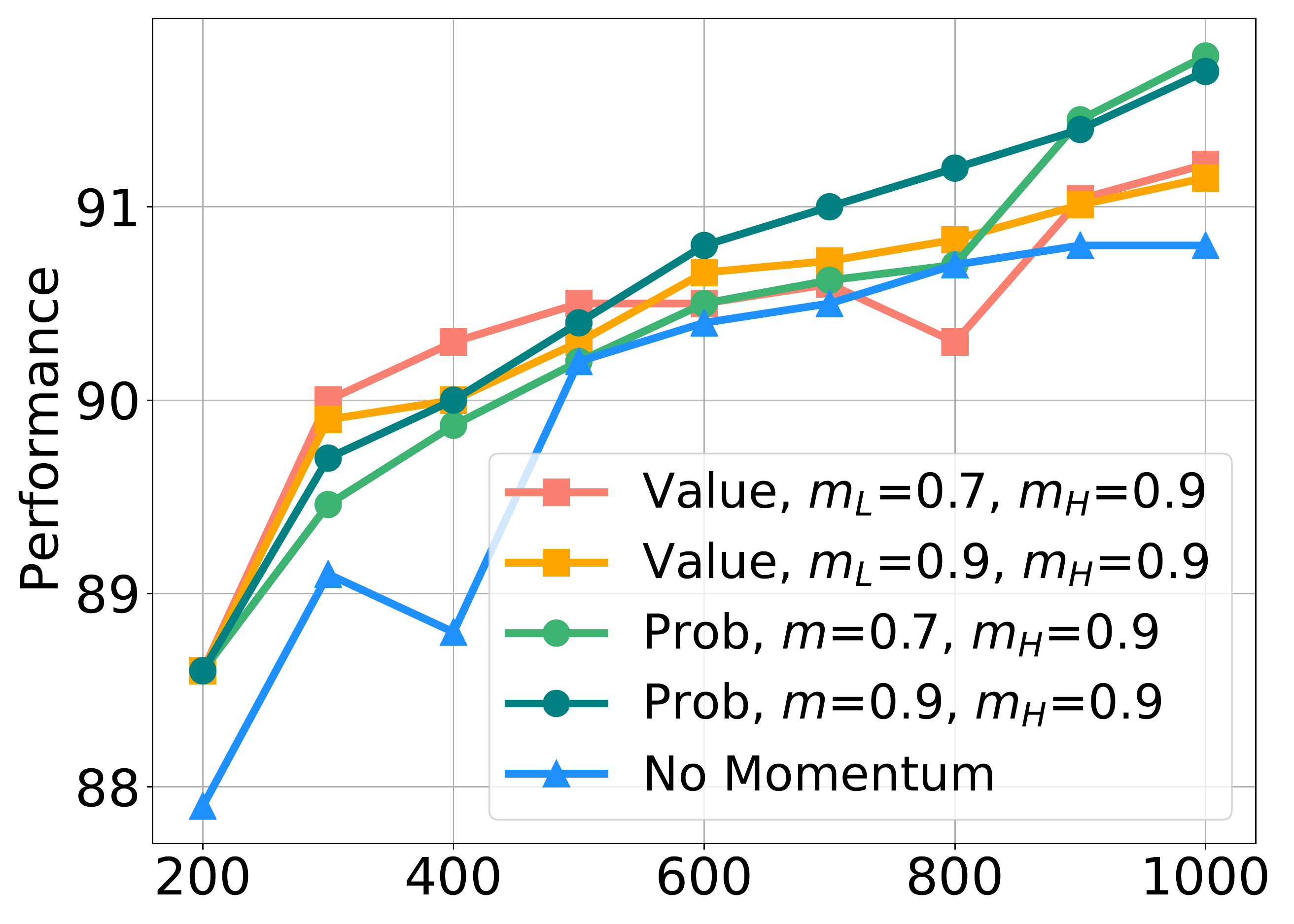}
            \label{fig:entropy_m}
        }\hspace{-2.3mm}
        \subfigure[CAL]{
            \includegraphics[width=0.19\textwidth]{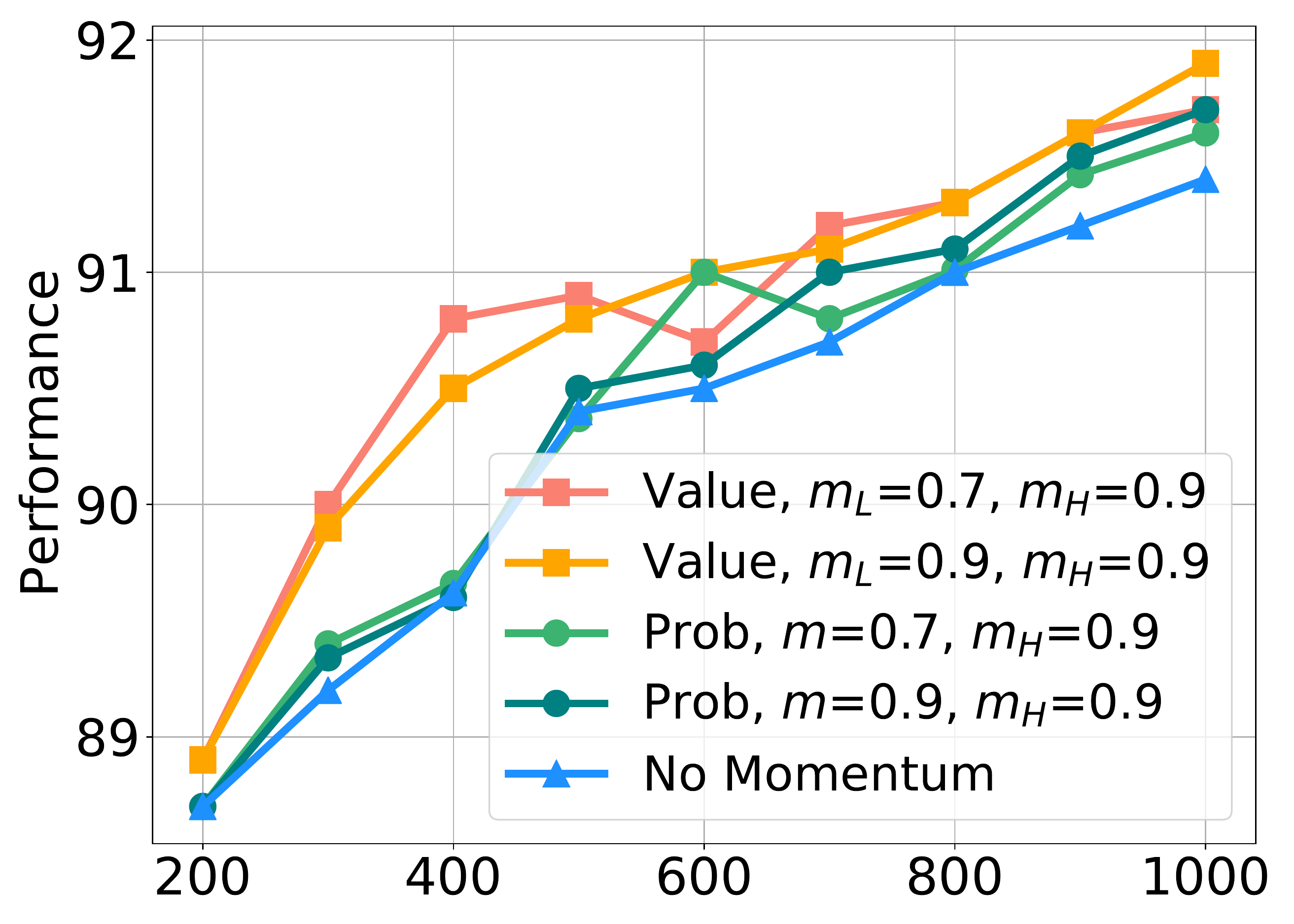}
            \label{fig:cal_m}
        }
            \vspace{-2ex}
        \caption{Parameter study. Note the effect of different $m_L$ and $m_H$ is conducted on AG News dataset.}\label{fig:ablation}
         \vspace{-3ex}
\end{figure*}

\begin{figure}[t]
    \vspace{-2mm}
        \centering
        \subfigure[Combining w/ AL Methods]{
    \includegraphics[width=0.23\textwidth]{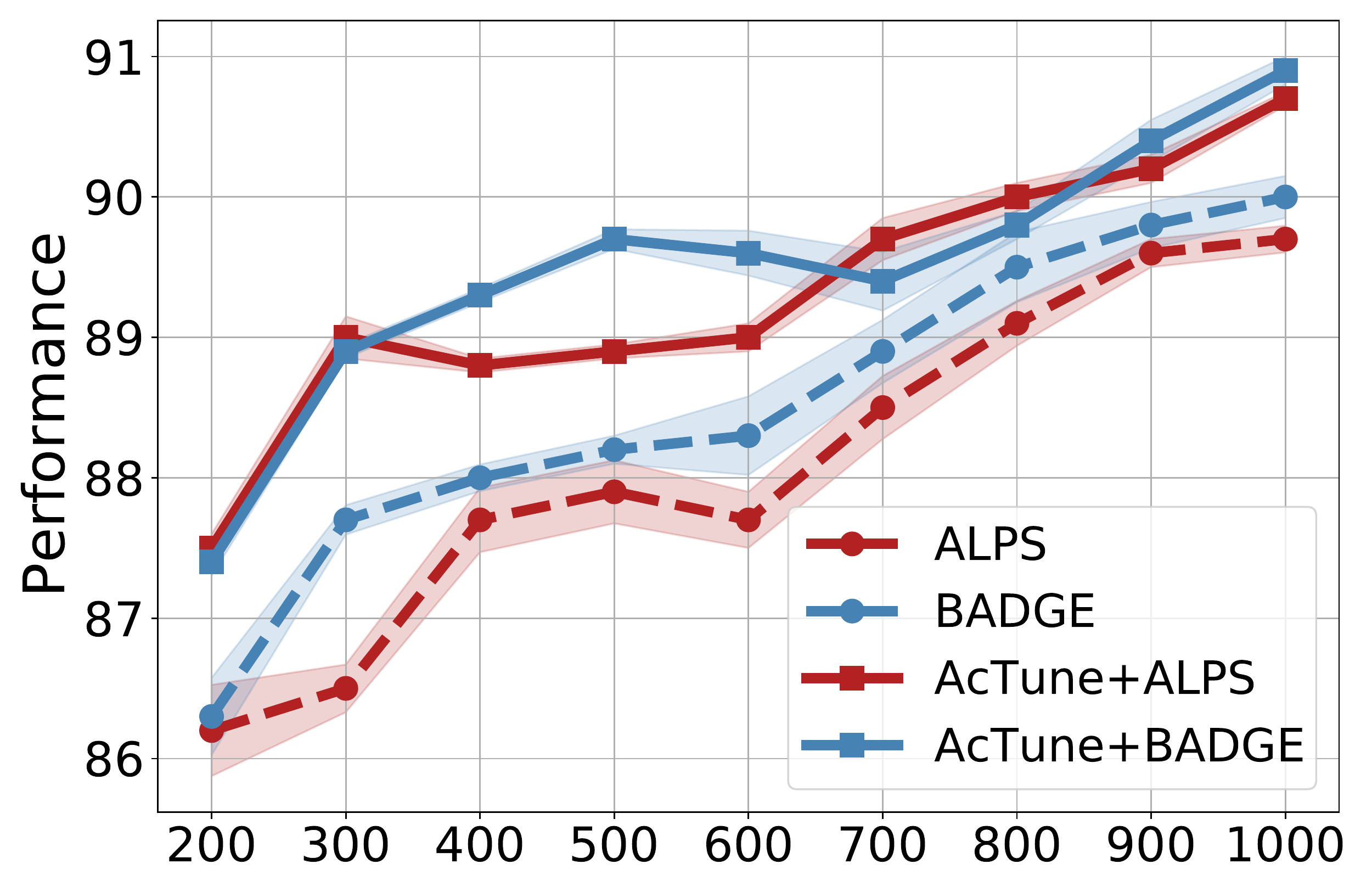}
            \label{fig:ablation_badge}
        }\hspace{-3mm}
        \subfigure[Ablation Study]{
            \includegraphics[width=0.23\textwidth]{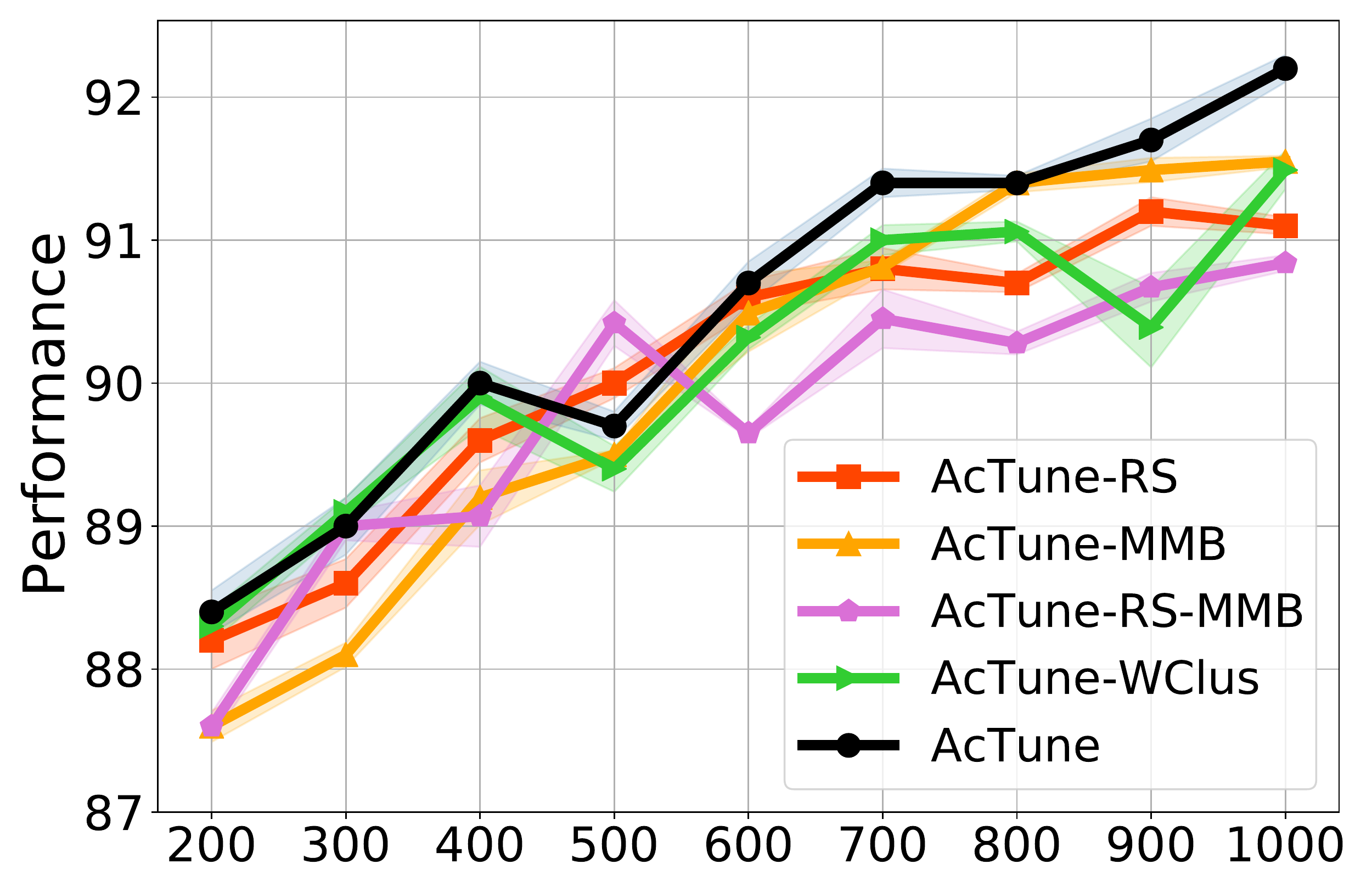}
            \label{fig:ablation_components}
        }
            \vspace{-2ex}
        \caption{Results of {\ours} with different AL methods (SST-2), ablation study (SST-2 with {\ours}+Entropy).} \label{fig:ablation}
         \vspace{-2ex}
\end{figure}
\subsection{Weakly-supervised Learning}
\label{sec:weak}
{\ours} can be naturally used for weakly-supervised classification, where $\cX_l$ is a set of noisy labels derived from linguistic patterns or rules. 
Since the initial label set is noisy, the model trained with Eq.~\eqref{eq:init} can  overfit the label noise, and we can actively query labeled data to refine the model.
We conduct experiments on the TREC and Chemprot dataset\footnote{Details for labeling functions are in \citet{zhang2021wrench}.}, where we first use Snorkel~\cite{ratner2017snorkel} to obtain weak label set $\cX_l$, then fine-tune the pre-trained LM $f(\theta^{(0)})$ on $\cX_l$. After that, we adopt {\ours} for active self-training.

Fig.~\ref{fig:weak} shows the results of these two datasets\footnote{We omit AL methods since they perform worse than SSAL methods on these datasets in general.}. 
When combining {\ours} with CAL, the performance is unsatisfactory. 
We believe it is because CAL requires clean labels to calculate uncertainties. When labels are inaccurate, it will prevent {\ours} from querying informative samples.
In contrast, {\ours} achieves the best performance over baselines when using Entropy as the uncertainty measure. 
The performance gain is more notable on the TREC dataset, where we achieve 96.68\% accuracy, close to the fully supervised performance (96.80\%) with only $\sim$6\% of clean labels.

\subsection{Combination with Other AL Methods}
Fig.~\ref{fig:ablation_badge} demonstrates the performance of {\ours} combined with other AL methods (e.g. BADGE, ALPS) on SST-2 dataset. 
It is clear that even if the AL methods are not uncertainty-based (e.g. BADGE), 
when using the \emph{entropy} as an uncertainty measure to select pseudo-labeled data for self-training, 
{\ours} can further boost the performance. 
This indicates that {\ours} is a general active self-training approach
, as it can serve as an efficient plug-in module for existing AL methods.

\subsection{Ablation and Hyperparameter Study}
\label{sec:ablation}
\textbf{The Effect of Different Components in {\ours}.} We inspect different components of {\ours}, including the region-sampling (RS), momentum-based memory bank (MMB), and weighted clustering (WClus)\footnote{For models w/o RS, we directly select samples with highest uncertainty during AL. 
For models w/o MMB, we only use the prediction from the current round for self-training.
For models w/o WClus, we cluster data with vanilla K-Means.}.
Experimental results (Fig.~\ref{fig:ablation_components}) shows that all the three components contribute to the final performance, as removing any of them hurts the classification accuracy. 
Also, we find that when removing MMB, the performance hurts most in the beginning rounds, which indicates that MMB effectively suppresses label noise when the model's capacity is weak. Conversely, removing WClus hurts the performance on later rounds, as it enables the model to select most informative samples.

\noindent \textbf{Hyperparameter Study.} We study two hyperparameters, namely $\beta$ and $K$ used in querying labels. Figure~\ref{fig:param_beta} and \ref{fig:param_K} show the results. In general, the model is insensitive to $\beta$ as the performance difference is less than 0.6\%. The model cannot perform well with smaller $K$ since it cannot pinpoint to high-uncertainty regions. For larger $K$, the performance also drops as some of the high-uncertainty regions can be outliers and sampling from them would hurt the model performance~\cite{karamcheti2021mind}.


\noindent \textbf{A Closer Look at the Momentum-based Memory Bank.}
To examine the role of MMB, we show the overall accuracy of pseudo-labels on AG News dataset in Fig.~\ref{fig:acc_pseudo}. 
From the result, it is clear that the momentum-based memory bank can stabilize the active self-training process, as the accuracy of pseudo labels increases around 1\%, especially in later rounds. 
Fig~\ref{fig:entropy_m} and \ref{fig:cal_m} illustrates the model performance with different $m_L$ and $m_H$. Overall, we find that our model is robust to different choices as {\ours} outperform the baseline without momentum update consistently. 
Moreover, we find that the larger $m_H$ will generally lead to better performance in later rounds. 
This is mainly because in later rounds, the model's prediction is more reliable. Conversely, at the beginning of the training, the model's prediction might be oscillating on unlabeled data. 
In this case, using a smaller $m_L$ will favor samples with consistent predictions to improve the robustness of active self-training.

Another finding is that for different AL methods, the optimal memory bank can be different. For Entropy, probability-based memory bank leads to a better result, while for CAL, simple aggregating over uncertainty score achieves better performance. This is mainly because the method used in CAL is more complicated, and using probability-based memory bank may hurt the uncertainty calculation.

\subsection{Case Study}
\begin{figure*}[t]
    \vspace{-2mm}
        \centering
        \subfigure[AG News, Entropy]{
            \includegraphics[width=0.42\textwidth]{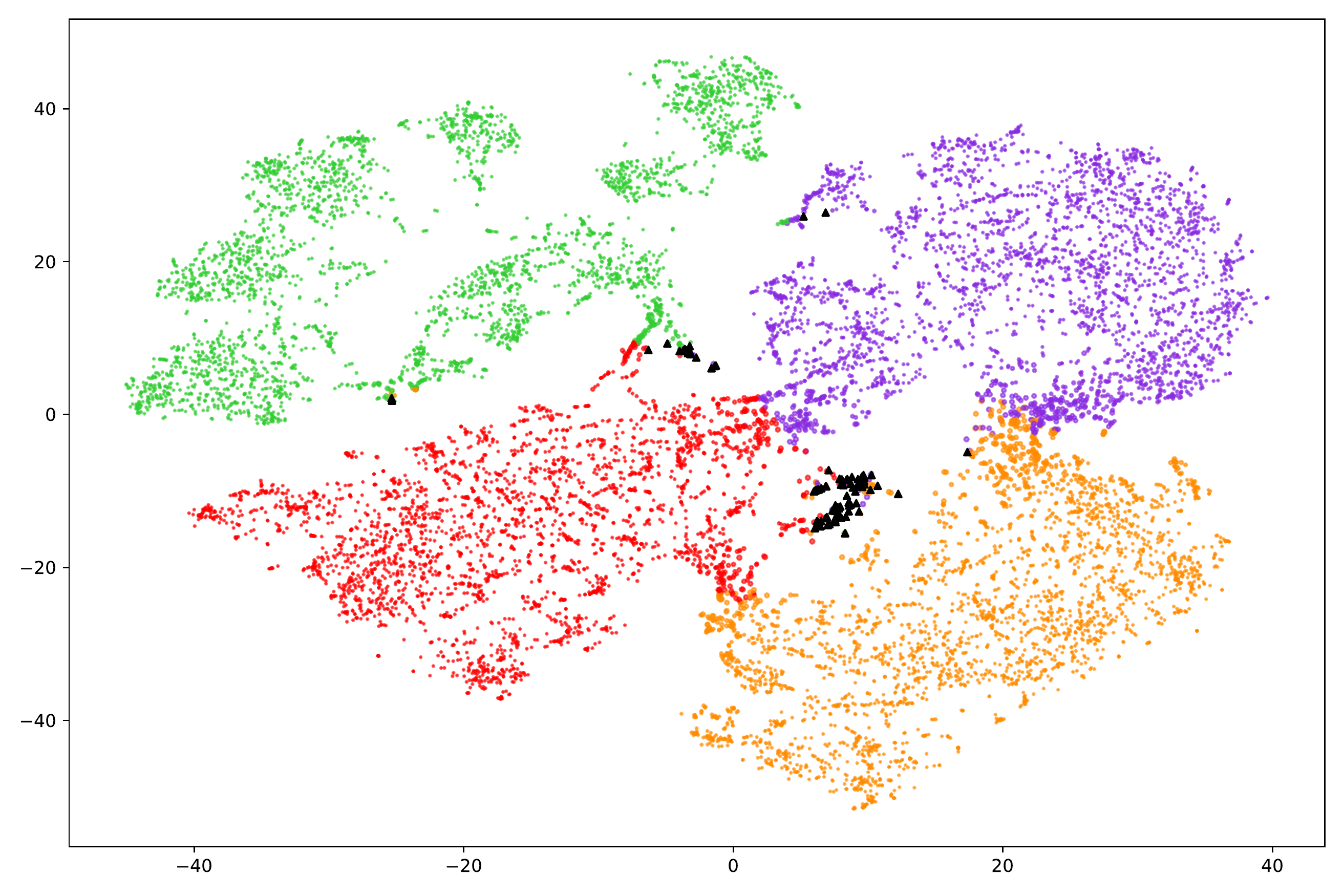}
        }
        \subfigure[AG News, {\ours} w/ Entropy]{
            \includegraphics[width=0.42\textwidth]{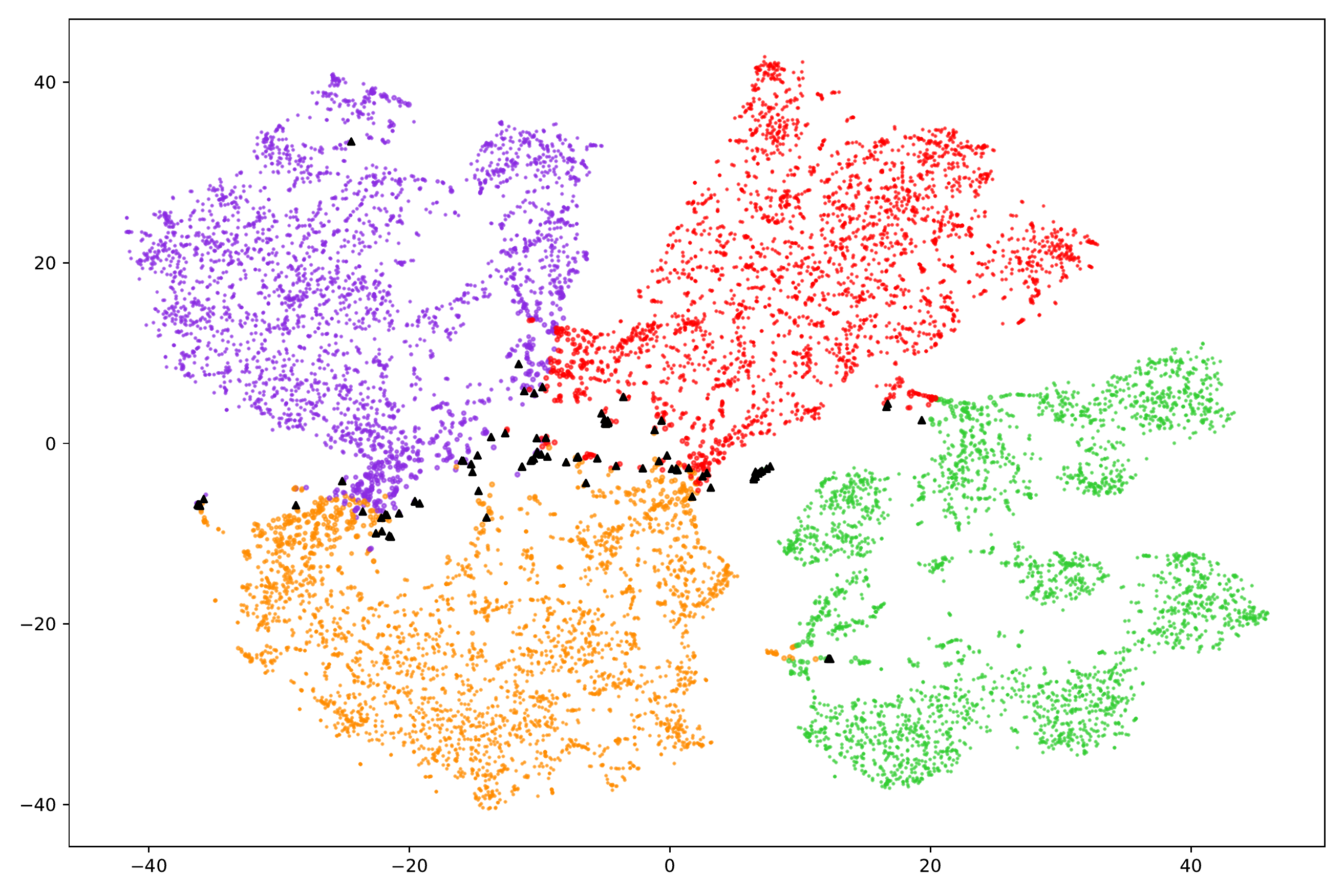}
        }
        \subfigure[AG News, CAL]{
            \includegraphics[width=0.42\textwidth]{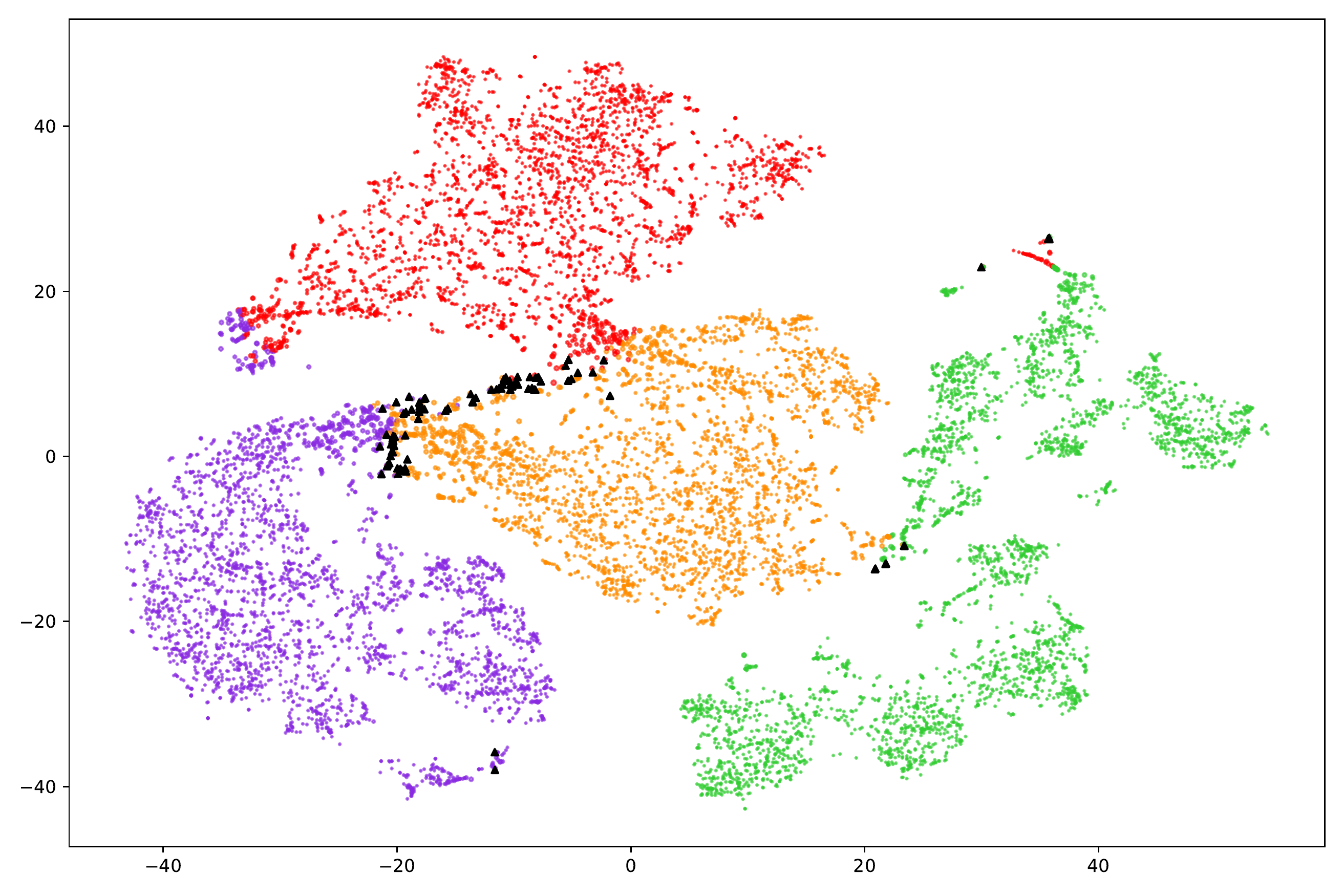}
        }
        \subfigure[AG News, {\ours} w/ CAL]{
    \includegraphics[width=0.42\textwidth]{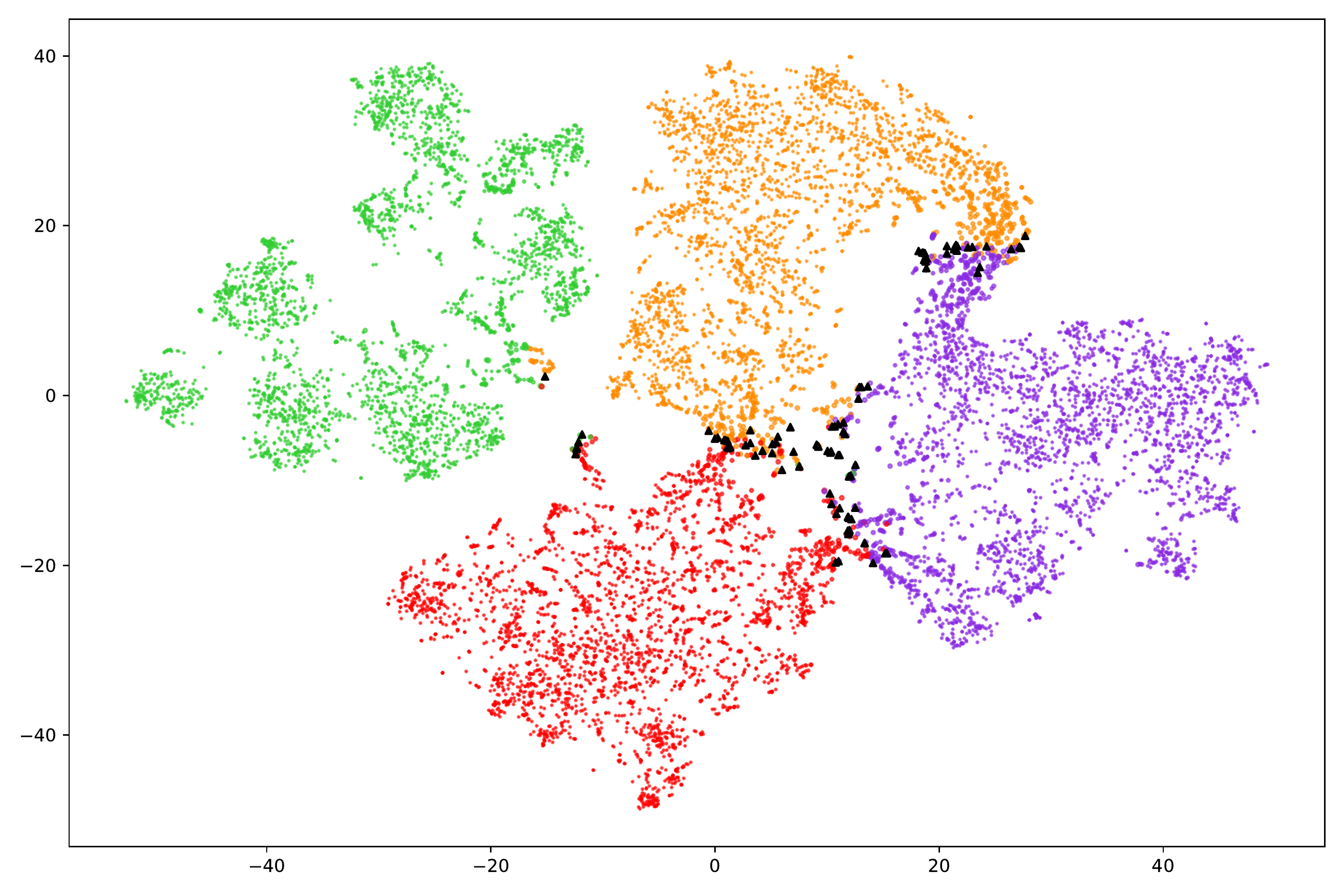}
        }
        \caption{Visualization of the querying strategy of {\ours}. Black dots stand for the queried data points. Different colors indicates different categories.}\label{fig:case_study}
         \vspace{-1ex}
\end{figure*}

We give an example of our querying strategy on AG News dataset for the 1st round of active self-training process in  figure~\ref{fig:case_study}. Note that we use t-SNE algorithm~\cite{tsne} for dimension reduction, and the black triangle stands for the queried samples while other circles stands for the unlabeled data. 
We can see that the existing uncertainty-based methods such as Entropy and CAL, are suffered from the issue of limited diversity. However, when combined with {\ours}, the diversity is much improved. Such results, compared with the main results in figure~\ref{fig:main}, demonstrate the efficacy of {\ours} empirically.

\section{Related Work}
\vspace{-2mm}
\textbf{Active Learning.} 
Active learning has been widely applied to various NLP tasks~\citep{yuan-etal-2020-cold,shelmano2021active,karamcheti2021mind}.
So far, AL methods can be categorized into uncertainty-based methods~\citep{gal2017deep,balm,margatina2021active}, diversity-based methods~\citep{ru2020active,coreset} and hybrid methods~\cite{yuan-etal-2020-cold,badge}. 
\citet{ein-etal-2020-active} offer an empirical study of active learning with PLMs. 
Very recently, there are also several works attempted to query labeling functions for weakly-supervised learning~\cite{boecking2020interactive,hsieh2022nemo,zhang2022prboost}.
In our study, we leverage the power of unlabeled instances via self-training to further promote the performance of AL.

\noindent  \textbf{Semi-supervised Active Learning (SSAL).} 
\citet{gao2020consistency,Guo_2021_ICCV} design query strategies for specific semi-supervised methods, 
\citet{zhang2020seqmix,jiang-etal-2020-camouflaged} combine active learning with data augmentation  and  
\citet{tomanek2009semi,rottmann2018deep,simeoni2021rethinking} exploit the most-certain samples from the unlabeled with pseudo-labeling to augment the training set. 
So far, most of the SSAL approaches are designed for CV domain and it remains unknown how this paradigm performs with PLMs on NLP tasks. In contrast, we propose {\ours} to effectively leverage unlabeled data during finetuing PLMs for NLP tasks. 


\noindent \textbf{Self-training.}
Self-training is one of the earliest and simplest approaches to semi-supervised learning~\cite{lee2013pseudo}.
It first generates pseudo labels for high-confidence samples, then fits a new model on pseudo labeled data to improve the generalization ability. 
However, it is known to be vulnerable to error propagation~\cite{arazo2020pseudo,rizve2021in,zuo2021self}.
To alleviate this, we adopt a simple momentum-based method to select high confidence samples, effectively reducing the pseudo labels noise for active learning. 
Note that although \citet{mukherjee2020uncertainty,rizve2021in} also leverage uncertainty information for self-training, their focus is on developing better self-training methods, while we aim to jointly query high-uncertainty samples and generate pseudo-labels for low-uncertainty samples. 
The experiments in Sec. \ref{sec:exp} show that with appropriate querying methods, {\ours} can further improve the performance of self-training.

\section{Conclusion and Discussion}
\vspace{-1mm}

In this paper, we develop \ours, a general active self-training framework for enhancing both label efficiency and model performance in fine-tuning pre-trained language models (PLMs). 
We propose a region-aware sampling approach to guarantee both the uncertainty the diversity for querying labels.
To combat the label noise propagation issue, we design a momentum-based memory bank to effectively utilize the model predictions for  preceding AL rounds.
Empirical results on 6 public text classification benchmarks suggest the superiority of {\ours} to conventional active learning and semi-supervised active learning methods for fine-tuning PLMs with limited resources.

There are several directions to improve~{\ours}. First, since our focus is on fine-tuning pre-trained language models, we use the representation of \texttt{[CLS]} token for classification. 
In the future work, we can consider using prompt tuning~\cite{gao2021making,schick-schutze-2021-exploiting}, a more data-efficient method for adopting pre-trained language models on classification tasks to further promote the efficiency. 
Also, due to the computational resource constraints, we do not use larger pre-trained language models such as RoBERTa-large~\cite{liu2019roberta} which shown even better performance with only a few labels~\cite{du-etal-2021-self}.
Moreover, we can explore more advanced uncertainty estimation approach~\cite{kong2020sde} into {\ours} to further improve the performance. 
Last, apart from the text classification task, we can also extend our work into other tasks such as sequence labeling and natural language inference (NLI).

\section*{Acknowledgements}

We thank the anonymous reviewers for their feedback. This work was supported in part by NSF IIS-2008334, IIS-2106961, CAREER IIS-2144338, and ONR MURI N00014-17-1-2656.

\bibliography{anthology}
\bibliographystyle{acl_natbib}

\clearpage
\appendix
\section{Datasets Details}
\subsection{Data Source}

The seven benchmarks in our experiments are all publicly available. Below are the links to downloadable versions of these datasets.

\noindent $\diamond$ \textit{SST-2}: We use the datasets from  \url{https://huggingface.co/datasets/glue}.

\noindent $\diamond$ \textit{AGNews}: We use the datasets from \url{https://huggingface.co/datasets/ag_news}.

\noindent $\diamond$ \textit{Pubmed-RCT}: Dataset is available at \url{https://github.com/Franck-Dernoncourt/pubmed-rct}.

\noindent $\diamond$ \textit{DBPedia}: Dataset is available at \url{https://huggingface.co/datasets/dbpedia_14}.

For two weakly-supervised classification tasks, we use the data from WRENCH benchmark~\cite{zhang2021wrench}. 

\noindent $\diamond$ \textit{TREC}: Dataset is available at \url{https://drive.google.com/drive/u/1/folders/1v55IKG2JN9fMtKJWU48B_5_DcPWGnpTq}.

\noindent $\diamond$ \textit{ChemProt}: The raw dataset is available at \url{http://www.cbs.dtu.dk/services/ChemProt/ChemProt-2.0/}. The preprocessed dataset is available at \url{https://drive.google.com/drive/u/1/folders/1v55IKG2JN9fMtKJWU48B_5_DcPWGnpTq}.

\subsection{Train/Test Split}
For all the datasets, we use the original train/dev/test split from the web. 
To keep the size of the development set small, we randomly sample 1000 data for \textit{SST-2, AGNews, Pubmed-RCT, DBPedia} and randomly sample 500 samples for \textit{TREC, ChemProt}.

\section{Details on Implementation and Experiment Setups}
\label{appendix:exp}

\subsection{Computing Infrastructure}
\textit{System}: Ubuntu 18.04.3 LTS; Python 3.6; Pytorch 1.6. \\
\textit{CPU}: Intel(R) Core(TM) i7-5930K CPU @ 3.50GHz. \\
\textit{GPU}: NVIDIA 2080Ti. \\

\begin{table*}[t]
	\begin{center}
		\begin{tabular}{c|c|c|c|c|c|c}
			\toprule 
			\bf Hyper-parameter &\bf SST-2 & \bf  AG News & \bf Pubmed & \bf DBPedia & \bf TREC  & \bf Chemprot \\ \midrule 
			Dropout Ratio & \multicolumn{6}{@{\hskip1pt}c@{\hskip1pt}}{0.1}  \\ \hline
			 Maximum Tokens  & 32 & 96 & 96 & 64 & 64 & 128  \\ \hline 
			 Batch Size for $\cX_l$ & \multicolumn{6}{@{\hskip1pt}c@{\hskip1pt}}{8}   \\ \hline
            Batch Size for $\cX_u$ in Self-training  & 32 & 48 & 48 & 32 & 16 & 24   \\ \hline
			 Weight Decay & \multicolumn{6}{@{\hskip1pt}c@{\hskip1pt}}{$10^{-8}$}  \\ \hline
			 Learning Rate & \multicolumn{6}{@{\hskip1pt}c@{\hskip1pt}}{$2\times 10^{-5}$}\\ \hline
			 $\beta$  & \multicolumn{6}{@{\hskip1pt}c@{\hskip1pt}}{0.5}\\ \hline
			 $M$ & 25 & 30 & 30 & 40 & 40 & 40 \\ \hline
             $K$ & 5 & \multicolumn{5}{@{\hskip1pt}c@{\hskip1pt}}{10} \\ \hline
             $\gamma$ &  0.7 & \multicolumn{5}{@{\hskip1pt}c@{\hskip1pt}}{0.6}  \\ \hline
			 $m_L$ & 0.8 & 0.9 & 0.7 & 0.8 & 0.8 & 0.8 \\ \hline
			 $m_H$ & 0.9 & 0.9 & 0.8 & 0.9 & 0.9 & 1.0 \\ \hline
             $\lambda$ & \multicolumn{6}{@{\hskip1pt}c@{\hskip1pt}}{1}  \\
			 \bottomrule
		\end{tabular}
	\end{center}
	\vspace{-1ex}
	\caption{Hyper-parameter configurations. Note that we only keep certain number of tokens.}
	\vspace{-1ex}
	\label{tab:hyperparameter}
\end{table*}

\begin{table}[h]
\centering
\begin{adjustbox}{max width=0.46\textwidth}
\begin{tabular}{|l|c|c|}
\hline
\multicolumn{1}{|c|}{\multirow{2}{*}{Method}} & \multicolumn{2}{c|}{Dataset}                           \\ \cline{2-3} 
\multicolumn{1}{|c|}{}                        & \multicolumn{1}{c|}{Pubmed}          & DBPedia         \\ \hline
Finetune (Random)                             & {\textless{}0.1s} & \textless{}0.1s \\ \hline
Entropy~\cite{holub2008entropy}                  & {461s}            & 646s            \\ \hline
BALD~\cite{gal2017deep}               & {4595s}           & 6451s           \\ \hline
ALPS~\cite{yuan-etal-2020-cold}                                          & {488s}            & 677s            \\ \hline
BADGE~\cite{badge}                                   & {554s}            & 1140s           \\ \hline
CAL~\cite{margatina2021active}                                           & {493s}            & 688s            \\ \hline
REVIVAL~\cite{Guo_2021_ICCV}                                       & {3240s}           & OOM             \\ \hline
{\ours} + Entropy                & 477s   &     733s            \\ \hline
\ \ \ w/ RS for Active Learning                 & {15.8s}           & 44.9s           \\ \hline
\ \ \ w/ MMB for Self-training                                 & {0.12s}           & 0.18s           \\ \hline
{\ours} + CAL                    & 510s                &       735s          \\ \hline
\ \ \ w/ RS for Active Learning                    & {16.6s}           & 46.4s           \\ \hline
\ \ \ w/ MMB for Self-training                           & {0.12s}           & 0.18s           \\ \hline
\end{tabular}
\end{adjustbox}
\vspace{-1ex}
\caption{The running time of different baselines. Note that for ASST, CEAL and BASS, they directly use existing active learning methods so we do not list the running time here.}
\vspace{-3ex}
	\label{tab:time}
\end{table}

\subsection{Number of Parameters}
{\ours} and all baselines use Roberta-base~\cite{liu2019roberta} with a task-specific classification head on the top as the backbone, which contains 125M trainable parameters. We do not introduce any other parameters in our experiments.

\subsection{Experiment Setups}
Following~\cite{ein-etal-2020-active,yuan-etal-2020-cold,margatina2021active}, all of our methods and baselines are run with 3 different random seed and the result is based on the average performance on them. This indeed creates $4$ (the number of datasets) $\times$ $3$ (the number of random seeds) $\times$ $11$ (the number of methods) $\times$ $10$ (the number of fine-tuning rounds in AL) = $1320$ experiments for fine-tuning PLMs, which is almost the limit of our computational resources, not to mention additional experiments on weakly-supervised text classification as well as different hyper-parameter tuning.
We have show both the mean and the standard deviation of the performance in our experiment sections. All the results have passed a paired t-test with $p<0.05$~\cite{dror2018hitchhiker}.











\subsection{Hyper-parameters for General Experiments}
\label{appendix:hyper}
We use AdamW  as the optimizer, and the learning rate is chosen from $1\times 10^{-5}, 2\times 10^{-5}\}$. 
A linear learning rate decay schedule with warm-up $0.1$ is used, and the number of training epochs is $15$ for fine-tuning. For active self-training \& SSAL baselines, we tune the model with 2000 steps, and evaluate the performance on the development set in every 50 steps. Finally, we use the model with best performance on the development set for testing.

\subsection{Hyper-parameters for {\ours}}
\label{appendix:ours}


 Although {\ours} introduces several
hyper-parameters including $K$, $M$, $m_L$, $m_H$, $\beta, \gamma$, $\lambda$, most of them are keep fixed  during our experiments, thus it does not require heavy hyper-parameter tuning. 
All results are reported as the average over three runs.

In our experiments, we keep $\beta=0.5$, $\lambda=1$ for all datasets. 
For other parameters, we use a grid search to find the optimal setting for each datasets.
Specifically, we search $\gamma$ from $[0.5, 0.6, 0.7]$, $m_L$ from $[0.6, 0.7, 0.8]$, $m_H$ from $[0.8, 0.9, 1]$. 
For {\ours} with Entropy, we use probability based aggregation and for  {\ours} with CAL, we use value based aggregation by default.



 \section{Runtime Analysis}

\label{sec:time}
Table~\ref{tab:time} shows the time in one active learning round of {\ours} and  baselines. Here we highlight that the additional time for region-aware sampling and momentum-based memory bank is \emph{rather small} compared with the inference time.
Also, we find that BALD and REVIVAL are not so efficient. For BALD, it needs to infer the uncertainty of the model by passing the data to model with multitple times. Such an operation will make the total inference time for PLMs very long.
For REVIVAL, we find that calculating the adversarial gradient needs extra forward passes and backward passes, which could be time-consuming for PLMs with millions of parameters\footnote{The original model is proposed with CV tasks and they use ResNet-18 as the backbone which only contains 11M parameters (around 10\% of the parameters of Roberta-base).}.


\end{document}